\documentclass[final,1p,times]{elsarticle}
\usepackage{float}
\usepackage{caption}
\usepackage{amssymb}
\usepackage{endnotes}
\usepackage{graphicx}
\usepackage{amsmath} 
\usepackage{algorithm}
\usepackage{rotating}
\usepackage{multirow}
\usepackage{adjustbox}
\usepackage{algpseudocode}
\usepackage[table]{xcolor}
\usepackage{booktabs}
\usepackage[colorlinks=true,linkcolor=black,citecolor=blue,urlcolor=blue,]{hyperref}
\usepackage[doipre={doi:~}]{uri}
\usepackage{tabularx, makecell}

\biboptions{sort&compress}

\journal{Advanced Engineering Informatics}
\begin{document}
\begin{frontmatter}

\title{Event-Based Vision Sensing and Its Application to Pedestrian Detection for Intelligent Transportation and Surveillance}

\author[label1]{Han Wang}
\author[label1]{Juntao Wu}
\author[label1,label4]{Jingyuan Bao}
\author[label3]{Min Liu}
\author[label1]{Yaoxiong Wang}
\author[label1]{Saiao Zhou}
\author[label1]{Yuman Nie \corref{cor1}}
\author[label2]{Yun Li \corref{cor2}}

\cortext[cor1]{Corresponding author: Yuman Nie (email: nie@iim.cas.cn).}
\cortext[cor2]{Corresponding author: Yun Li (email: Yun.Li@uestc.edu.cn). Yuman Nie and Yun Li contributed equally as corresponding authors.}

\affiliation[label1]{organization={Hefei Institutes of Physical Science},
            addressline={Chinese Academy of Sciences}, 
            city={Hefei},
            postcode={230031}, 
            state={Anhui},
            country={China}}

\affiliation[label2]{organization={Shenzhen Institute for Advanced Study},
            city={Shenzhen},
            postcode={518110}, 
            state={Guangdong},
            country={China}}

\affiliation[label3]{organization={DongWeiShiJue (Beijing) Technology Co},
            city={Beijing},
            postcode={100080}, 
            state={Beijing},
            country={China}}

\affiliation[label4]{organization={University of Science and Technology of China},
            city={Hefei},
            postcode={230031}, 
            state={Anhui},
            country={China}}

\begin{abstract}
Pedestrian detection in conventional frame-based imaging often suffers from limited temporal responsiveness and substantial data redundancy. Inspired by the biological retina, event-based vision sensing (EVS) offers ultra-low latency, high temporal resolution, wide dynamic range, and low power consumption, making it highly attractive for pedestrian perception in complex environments. This paper provides a comprehensive review of EVS and its application to pedestrian detection in intelligent transportation and surveillance scenarios. We first summarize the sensing principles, historical development, and key advantages of event-based vision in comparison with conventional frame-based imaging. We then review the major methodological components of event-based pedestrian detection, including sensing inputs, event representations, preprocessing strategies, feature extraction, detection models, datasets, and evaluation metrics. In addition, representative methods are comparatively analyzed in terms of temporal fidelity, detection accuracy, computational efficiency, and deployment complexity. Finally, we discuss the major open challenges in current EB-PD research, including benchmark standardization, event-native model design, multimodal fusion, and real-world deployment, and outline several promising directions for future development. This review aims to provide a structured and up-to-date reference for researchers working on event-based pedestrian perception and related intelligent vision systems.
\end{abstract}

\begin{graphicalabstract}
\includegraphics[width=\textwidth]{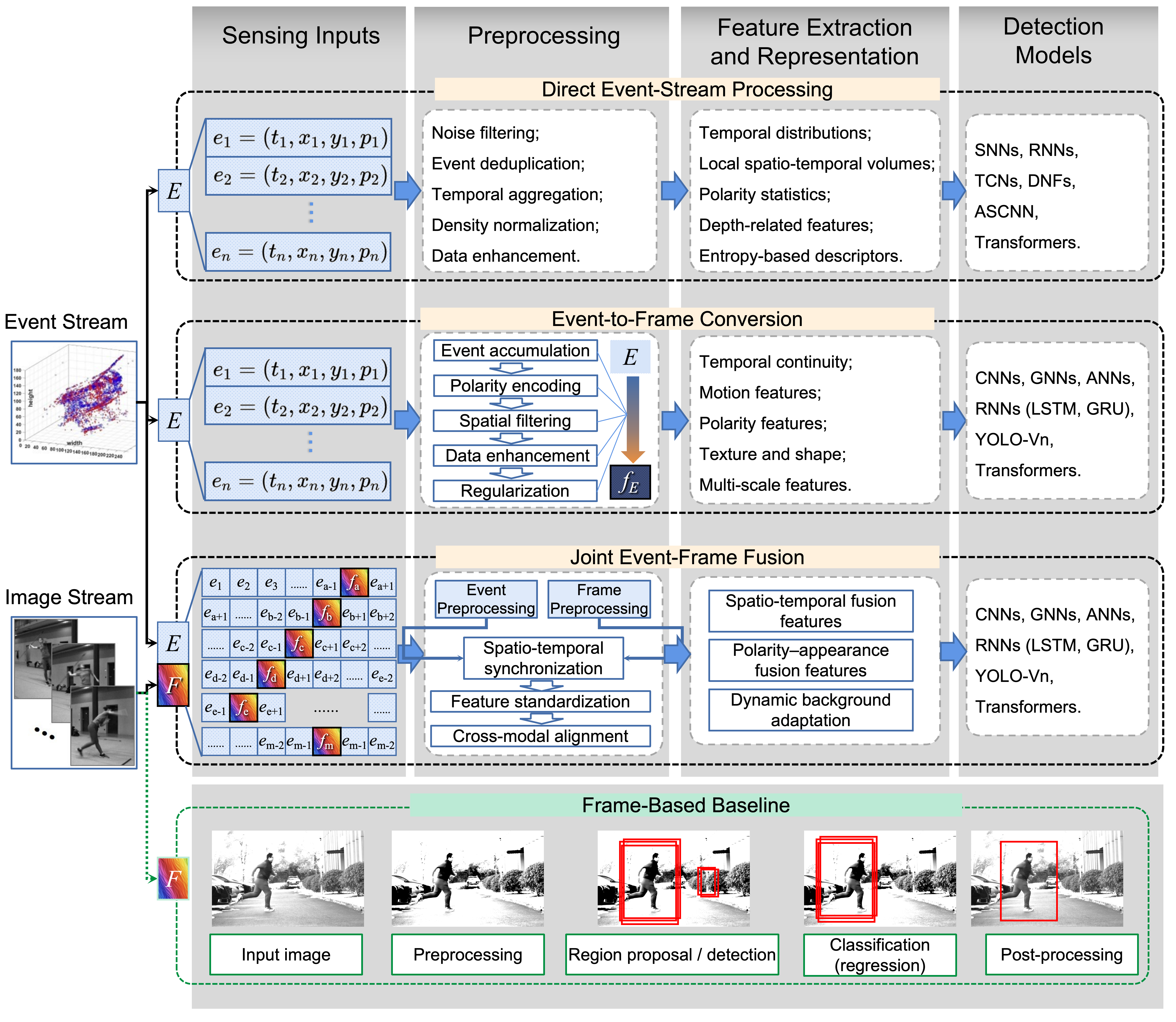}
\end{graphicalabstract}

\begin{highlights}
  \item A comprehensive review is presented on event-based vision sensing and its application to pedestrian detection in intelligent transportation and surveillance.

  \item Existing event-based pedestrian detection studies are structured around sensing inputs, preprocessing strategies, feature representations, detection models, datasets, and evaluation metrics.

  \item Representative methods are critically compared from the perspective of temporal fidelity, detection accuracy, computational efficiency, and deployment complexity.

  \item The review highlights key trade-offs in event-based pedestrian detection, including direct event streams versus event-to-frame conversion, SNN-based versus CNN-based processing, and short versus long temporal accumulation windows.

  \item Open challenges are discussed with emphasis on benchmark standardization, event-native perception models, multimodal learning, and deployment-oriented research.

\end{highlights}

\end{frontmatter}
\section{INTRODUCTION}
\label{sec1}

Pedestrian detection is a fundamental task in computer vision. It aims to accurately identify \cite{1pd,2pd,3pd,4pd} and localize \cite{5pd,6pd,7pd} pedestrians in images or image sequences. As a core component of Intelligent Transportation Systems (ITS) \cite{ITS1,ITS2,ITS3}, pedestrian detection plays an important role in active safety and autonomous driving \cite{AR1,AR2}. It is also closely related to pedestrian tracking \cite{genzong1,genzong2,genzong3,genzong4,genzong5,genzong6}, person re-identification \cite{chong1,chong2,chong3,chong4}, intelligent surveillance \cite{shipin1,shipin2,shipin3}, human behavior analysis \cite{renti1,renti2,renti3,renti4}, assisted driving \cite{fuzhu1,fuzhu2,fuzhu3}, public safety \cite{gonggong1,gonggong2,gonggong3}, and broader AI-driven perception systems \cite{rengong1,rengong2}. Existing pedestrian-detection methods include global feature-based approaches \cite{1pdmethod1,1pdmethod2,1pdmethod3,1pdmethod4,1pdmethod5}, body part-based methods \cite{2pdmethon1,2pdmethon2,2pdmethon3}, and stereo-vision-based techniques \cite{3pdmethon1,3pdmethon2,3pdmethon3,3pdmethon4,3pdmethon5}. Despite substantial progress, pedestrian detection remains challenging under difficult illumination, high-speed motion, occlusion, and resource-constrained deployment scenarios \cite{2pdmethon2,edge1}.

These challenges become even more pronounced in real-world traffic scenes. Pedestrian behavior is inherently difficult to predict because it depends on complex and dynamic factors, including motion state, intended destination, demographic attributes, and interactions with nearby vehicles and other pedestrians \cite{P1,P2,P3,P4,P5,P6,P7,P8,P9}. From an edge-computing perspective, pedestrian perception systems must also balance model complexity against computational efficiency and latency constraints \cite{edge1}. These considerations motivate the search for sensing and perception paradigms that can respond more rapidly and robustly to dynamic environments.

\begin{figure}[h]
    \centering
    \includegraphics[width=0.9\linewidth]{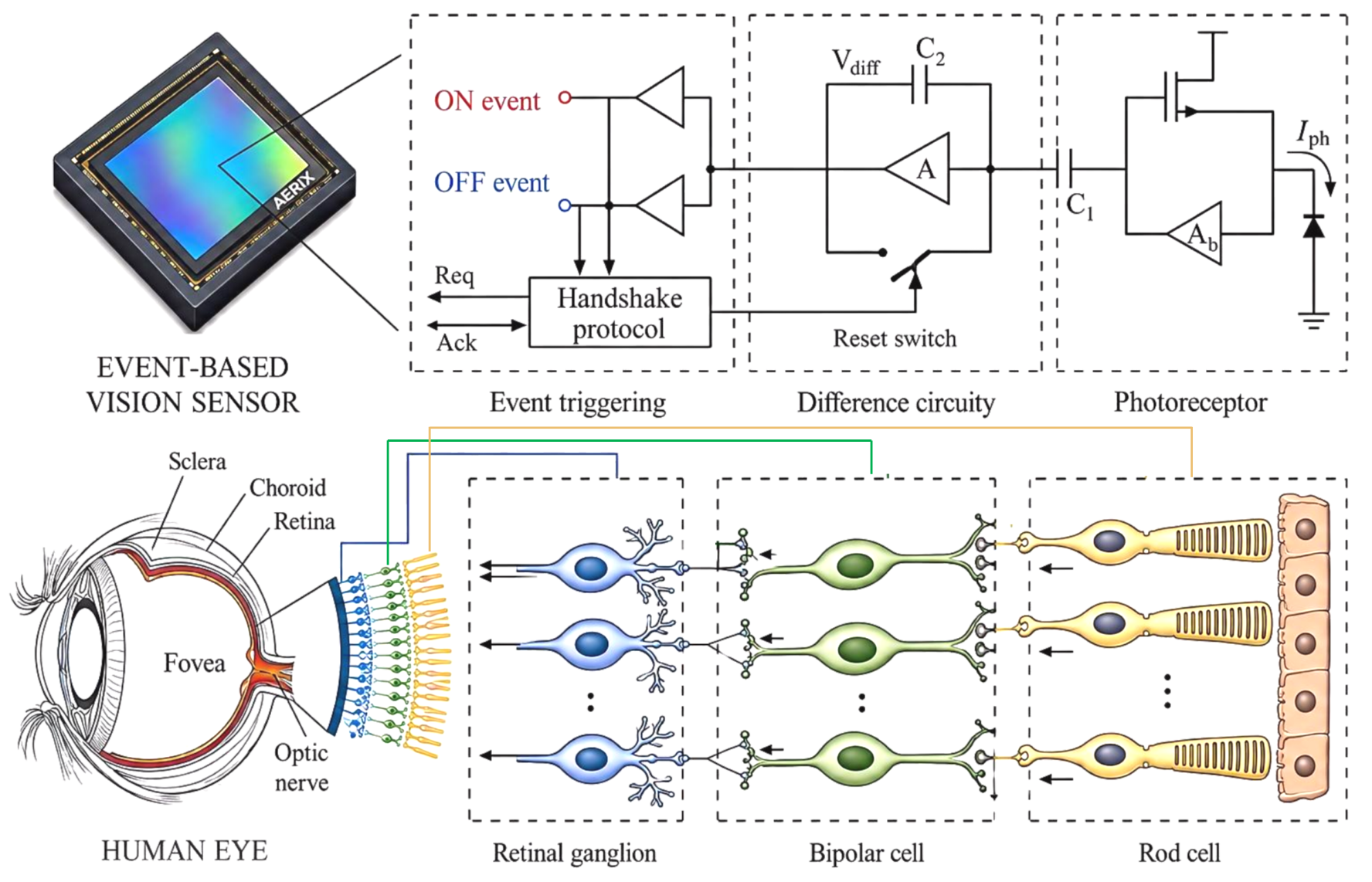}
    \caption{An event camera captures visual information in a retina-inspired manner. Each pixel operates independently and responds only to changes in light intensity. An ON event is triggered by an increase in brightness, whereas an OFF event is triggered by a decrease.}
    \label{fig:DCV}
\end{figure}

Event cameras are asynchronous visual sensors that represent a major shift in image acquisition \cite{survey1}. Their pixel design is inspired by the mammalian retina \cite{1survey}. Event-based vision sensing (EVS) \cite{EC1,EC2,EC3,EC4} and the Dynamic Active Pixel Vision Sensor (DAVIS) \cite{DVS4,DAVIS1} are among the most representative technologies in this family \cite{DAVIS2,DAVIS3}. The development of EVS can be traced back to the ``Silicon Retina'' introduced by Mahowald and colleagues in 1992 \cite{DVS1}. In 2006, Delbruck's group introduced the Dynamic Vision Sensor (DVS) \cite{EC2}, which marked a major milestone in the maturation of event-camera technology. This was followed by the introduction of DAVIS in 2013 \cite{DVS3} and the DAVIS346 color variant in 2017 \cite{DVS4}. In 2021, Prophesee and Sony released the EVK4 platform equipped with the IMX636 event sensor. In 2025, DVSense launched the DVSync series, representing a recent attempt to integrate high-resolution RGB and EVS sensing within a unified hardware system.

The biological inspiration of EVS lies in several functional mechanisms of the retina, as illustrated in Fig.~\ref{fig:DCV}. In the biological retina, ON and OFF bipolar cells respond to increasing and decreasing light intensity, respectively \cite{survey1,1survey}. Event cameras mimic this principle by generating events when local log-intensity changes exceed a threshold, thereby emphasizing contrast changes rather than recording redundant static information \cite{bi1,bi2,bi3}. In addition, retinal mechanisms related to contrast enhancement and information transmission have inspired the design of event-based sensing circuits, which encode scene dynamics with precise timestamps and polarity information \cite{bi1,bi3,bi4,DVS1,DVS3,survey2}.

Each event records the pixel location, timestamp, and polarity of a brightness change \cite{EC1,DVS3}. Because events are generated only when changes occur, event cameras offer several distinctive advantages over conventional frame-based imaging. First, they capture luminance variations with microsecond-level temporal resolution, making them highly suitable for fast motion without conventional motion blur \cite{survey1,survey22,tobi1,tobi2}. Second, their sparse output greatly reduces redundant data transmission and lowers power consumption \cite{survey1}. Third, their high dynamic range (HDR) enables robust sensing under challenging illumination conditions \cite{tobi3}. For pedestrian detection, these properties are particularly attractive because they support faster response to environmental changes \cite{AR1,PDS1}, more reliable operation under low light \cite{PDS2}, and improved perception of rapidly moving pedestrians \cite{PDS3}.

Against this background, this paper reviews the field of event-based pedestrian detection (EB-PD), with emphasis on applications in autonomous driving, intelligent transportation, and human-centered visual perception. We examine the available datasets, sensing settings, event representations, preprocessing pipelines, feature-extraction strategies, and detection models used in this domain. Existing methods are organized according to their sensing and representation paradigms, including direct event-stream processing, event-to-frame conversion, and joint event-frame fusion, and their strengths and limitations are analyzed accordingly.

The primary contributions of this review are threefold. First, we provide a structured analysis of the EB-PD literature and organize existing methods into three representative processing paradigms: direct event-stream processing, event-to-frame conversion, and joint event-frame fusion, while discussing their respective advantages and limitations. Second, we summarize representative public datasets and commonly used evaluation metrics, and we compare major EB-PD-related tasks and methods across widely used benchmarks. Third, we identify current research gaps, major practical challenges, and several promising directions for future development in event-based pedestrian detection.
\section{METHODOLOGY AND TAXONOMY} 
\subsection{Methodology}
Given the expansive definition \cite{survey1} of event-based sensors and the wide variety of related devices \cite{lack1}, a rigorous and unified system remains undeveloped \cite{lack2}. Before presenting the search process, we therefore clarify the scope of this review. In this paper, the term ``EBC'' is broadly used to denote vision sensors capable of capturing dynamic event signals, including technologies referred to as DVS, DAVIS, EVS, and related event-based cameras. Meanwhile, the EB-PD task considered here primarily focuses on pedestrian-related perception for intelligent transportation and active safety, while also encompassing scenario-specific tasks such as posture identification, static pedestrian detection, and road-target segmentation.

Our methodology for locating relevant literature involved both direct searching and the snowballing technique.
We utilized platforms such as IEEE Xplore digital library, Springer, ScienceDirect, ACM, and Google Scholar for direct search to include both scientific databases and open-access pre-prints.\footnote{The compiled documentation supporting our research is publicly available at \url{https://github.com/TristanWH/DVS4PD}.} Table~\ref{query} summarizes the search terms used on each platform and the number of records retrieved.
The search covered publications from 2014 to early 2026 and identified 353 relevant articles.

\newcolumntype{Y}{>{\centering\arraybackslash}p}
\begin{table*}[h]
  \caption{Query Overview}
  \centering
  \begin{tabular}{Y{0.1\linewidth} Y{0.08\linewidth} Y{0.70\linewidth}}
    \toprule
    \textbf{Database} & \textbf{\# Records} & \textbf{Query}\\
    \midrule
    IEEE & 72 & ``All Metadata'': (DVS AND Pedestrian Detection) OR (EVS AND Pedestrian Detection) OR (Event based AND Pedestrian
    Detection) OR (Neuromorphic Vision AND Pedestrian Detection) \\
    \addlinespace
    Elsevier & 19 & pub-date \textgreater 2014 and (DVS AND Pedestrian Detection) OR (EVS AND Pedestrian Detection) OR (Event based AND Pedestrian Detection) OR (Neuromorphic Vision AND Pedestrian Detection) \\
    \addlinespace
    Springer & 32 & (DVS AND Pedestrian Detection) OR (EVS AND Pedestrian Detection) OR (Event based AND Pedestrian Detection) OR (Neuromorphic Vision Pedestrian Detection) ``within 2014 - 2026'' \\
    \addlinespace
    ACM & 27 & [[All Metadata: ``DVS AND Pedestrian Detection'']] OR [[All Metadata: ``EVS AND Pedestrian Detection'']]  OR [[All Metadata: ``Event based AND Pedestrian Detection'']] OR [[All Metadata: ``Neuromorphic  vision AND Pedestrian Detection'']] AND [E-Publication Date: (01/01/2014 TO 12/31/2024)] \\
    \addlinespace
    Google Scholar & 193 & ``DVS AND Pedestrian Detection'' OR ``Event based Pedestrian Detection'' OR ``Neuromorphic Vision AND Pedestrian Detection'' OR ``EVS AND Pedestrian Detection'' custom range 2014 - present \\
    \bottomrule
  \end{tabular}
          \label{query}
\end{table*}

The primary objective of our inclusion and exclusion criteria was to identify recent and relevant peer-reviewed studies focused on the EB-PD task. Because search syntax differs across databases, the query expressions were adapted to each platform. The main requirement was that the search terms be explicitly mentioned within the articles' titles, abstracts, or keywords. In the initial screening, we applied the following preliminary exclusion criteria:
\begin{itemize}
    \item[] \textbf{F1.} Exclude review and survey papers.
    \item[] \textbf{F2.} Exclude publications that are neither conference nor journal articles, including master's and doctoral theses.
    \item[] \textbf{F3.} Exclude papers that mention EB-PD but do not focus on EB-PD tasks (such as image segmentation).
    \item[] \textbf{F4.} Exclude short papers and duplicates.
    \item[] \textbf{F5.} Exclude papers not written in English.
\end{itemize}

\begin{figure}[h]
        \centering
        \includegraphics[width=0.7\linewidth]{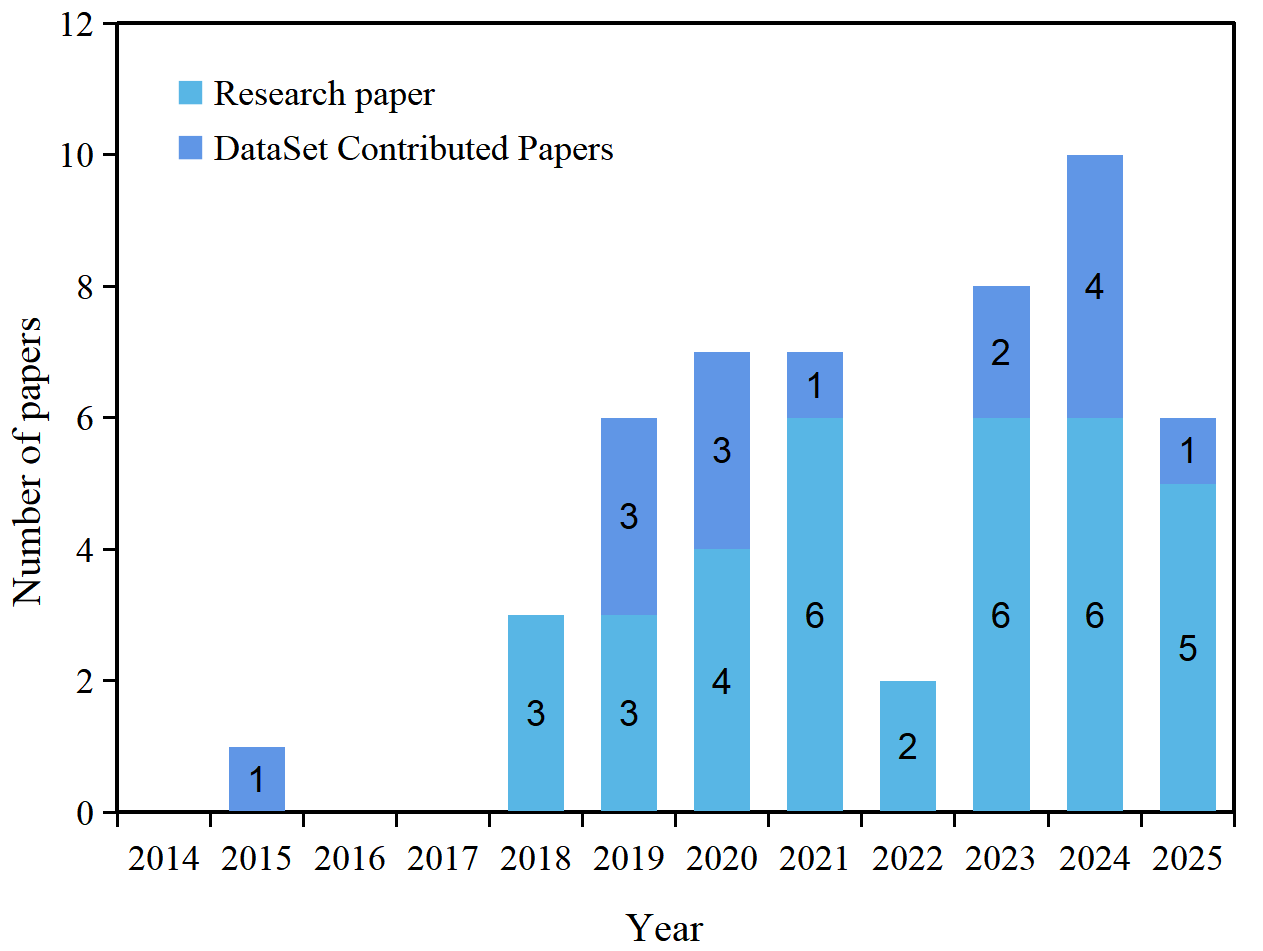}
        \caption{Number of high-quality research papers on EB-PD published over the past 12 years. The upward trend indicates growing research interest in EB-PD. Statistics are current as of 28 January 2026.}
        \label{fig:NAD}
    \end{figure}

Following this preliminary screening, we identified 50 papers across various platforms for inclusion. The distribution of these papers from 2014 to 2026 is detailed in Fig.~\ref{fig:NAD}.

Upon reviewing the corpus, it was determined that seven papers explicitly contribute to dataset development, focusing on the construction of comprehensive datasets for EB-PD.
A breakdown of the affiliations reveals that 31 papers were authored primarily by academics or researchers from institutions, while 19 originate from the industry, including notable contributions from corporations like Prophesee, Samsung, and Intel. 
These companies have advanced the field by sharing datasets derived from their state-of-the-art event cameras under diverse scenarios.
Moreover, 46.9\% of the included studies were published in recognized journals and conferences, including IEEE Conference on Computer Vision and Pattern Recognition Workshops (CVPRW),	IEEE Transactions on Pattern Analysis and Machine Intelligence (PAMI), IEEE Internet of Things Journal, IEEE Transactions on Information Forensics and Security, Neurorobotics, IEEE Access, Society of Photo-Optical Instrumentation Engineers (SPIE), and IEEE International Conference on Robotics and Automation (ICRA). 
These observations indicate sustained research interest in EB-PD and reflect the growing recognition of its technical challenges and application value.

\subsection{Taxonomy}

This section categorizes the extensive body of research into comprehensive domains, reflecting recent advancements and aligning with current technological trends.

\subsubsection{Data Input and Feature Extraction}
Research in EBC technology primarily focuses on the optimal utilization of the sparse and asynchronous data \cite{tay4, tay2} these sensors generate. 
Various methodologies for data capture are detailed, focusing on minimizing redundancy and improving the relevance of event signals through advanced filtering techniques \cite{tay1}. 
Feature extraction from EBC data involves transforming event-triggered intensity changes at specific pixels into formats suitable for pedestrian detection algorithms \cite{tay2}. 
This process often integrates temporal and spatial data to leverage the high temporal resolution of EBC, which is crucial for dynamic environments like urban traffic scenes \cite{tay3}.

\subsubsection{Detection Models and Model Integration}
Integrating EBC data with detection models poses unique challenges due to non-standard data formats.
Existing EB-PD studies can generally be organized into three sensing and representation paradigms \cite{HR-EC-PD, RTCM-GABP, MFE-FPD}. 
Converting event streams into frame-like representations and then processing them with conventional networks is common in both static- and dynamic-scene EB-PD. Meanwhile, researchers have also experimented with combining event streams and image frames as inputs \cite{zhen11, qiao2024spatio}. 
The potential for mining information in the event stream signals using mature frame-based deep learning models has been a major focus of recent research. 
Convolutional neural networks (CNNs) \cite{HR-EC-PD, E-YOLO-FPS, PoT-QY-PD, MC-EIF-PD, SSDE-RDVS, E-PD-DVS, MFE-FPD, GMVDT-NVS} have been adapted to handle the binary and sparse nature of EBC data, and modified to accommodate temporal dynamics absent in traditional video data.  
These models are designed to utilize the fine-grained temporal information from EBCs to improve detection accuracy and response times in pedestrian detection applications \cite{add1, add2}.

\subsubsection{Datasets, Evaluation Metrics, and Standards}
The development of dedicated datasets for EB-PD is a pressing research need, aiming to provide benchmarks that accurately reflect the operational challenges and capabilities of event-based sensors.
Open source datasets dedicated to EB-PD \cite{PEDRo2023, Prophesee2022, Neuromo2019, DVSOUTLAB2021} remain limited, but some researchers have conducted relevant experiments on other datasets \cite{Prophesee2022, SpikeEvent2023, DHP192019, NUAIRA2023, 1-M} that can be used for EB-PD. 
These datasets, although abundant in pedestrian data, do not specifically address unique scenarios or areas of particular interest for pedestrian detection.
Standardized dataset collection methods and clearly defined model evaluation metrics are imperative.
In addition to these datasets, evaluation metrics have been compiled to assess the performance of pedestrian detection systems under various operating conditions. 
These metrics address the accuracy, reliability, and computational efficiency of systems designed to process high-density event data.

    \begin{figure*}[!htb]
        \centering
        \includegraphics[width=0.9\textwidth]{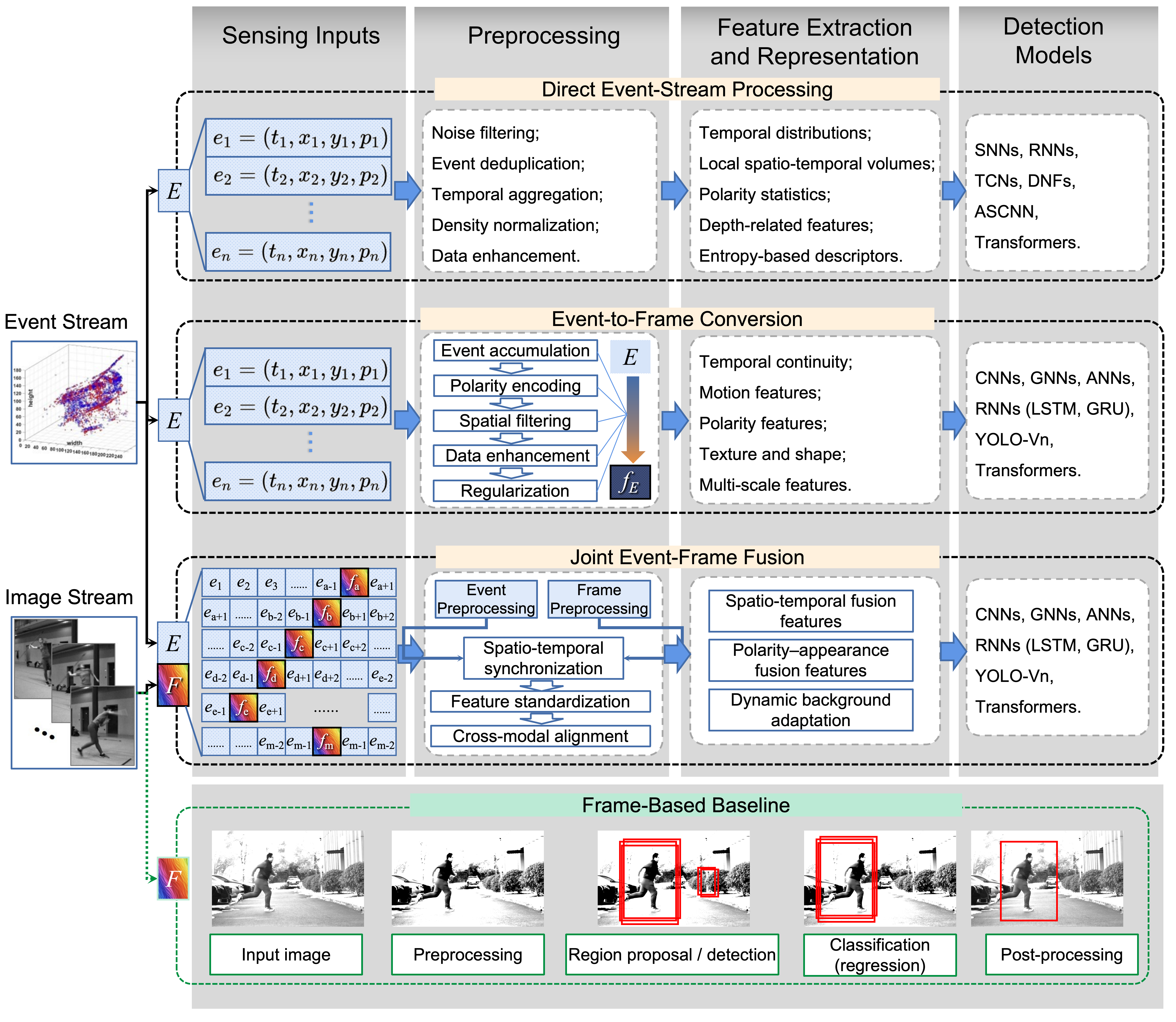}
        \caption{A generalized EB-PD pipeline consisting of four stages: sensing inputs, preprocessing, feature extraction and representation, and detection models. Existing methods can be broadly grouped into three paradigms: direct event-stream processing, event-to-frame conversion, and joint event-frame fusion. These paradigms differ mainly in how they preserve temporal information, exploit spatial structure, and balance event nativeness against compatibility with conventional vision models.}
        \label{fig:PIPELINE}
    \end{figure*}
    
\section{EVENT-BASED PEDESTRIAN DETECTION PIPELINE}
\label{sec:pedestrian_detection_pipeline}

Current EB-PD methods encompass a broad spectrum of paradigms, ranging from handcrafted approaches to deep learning-based and hybrid frameworks. In general, the EB-PD pipeline can be decomposed into four core stages: sensing inputs, preprocessing, feature extraction and representation, and detection models. Fig.~\ref{fig:PIPELINE} illustrates the overall workflow of these four stages. It should be noted, however, that not every method strictly follows this canonical pipeline. For example, the work in \cite{ApplicationOH} applies hierarchical clustering directly to event data, whereas \cite{EVSegNet2018} focuses on semantic segmentation based on event-frame signals. Nevertheless, from the perspective of methodological generalizability, the four-stage formulation remains a useful and sufficiently comprehensive abstraction for analyzing existing EB-PD methods.

\subsection{Sensing Inputs and Processing Paradigms}

The initial phase concerns the acquisition and organization of sensing inputs for downstream pedestrian detection. In existing EB-PD research, this stage can generally be categorized into three representative paradigms: direct event-stream processing, event-to-frame conversion, and joint event-frame fusion. The following subsections describe these three paradigms in detail.

\subsubsection{Direct Event-Stream Processing}

Direct Event-Stream Processing in EB-PD systems fully exploits the fine-grained temporal information captured by event cameras \cite{RTCM-GABP}. Unlike traditional cameras, which produce image frames at fixed intervals, event cameras output an asynchronous stream of events $\mathcal{E}$, where each event $e_i \in \mathcal{E}$ is represented as a tuple $(t_i, x_i, y_i, p_i)$:

\begin{itemize}
  \item $t_i$ denotes the timestamp at which a change in log intensity is detected;
  \item $(x_i, y_i)$ denote the spatial coordinates of the pixel where the event occurs;
  \item $p_i \in \{-1, +1\}$ denotes the polarity of the intensity variation.
\end{itemize}

This data stream is inherently sparse and asynchronous, and therefore requires dedicated preprocessing operations before it can be effectively exploited by pedestrian detection models \cite{step11, step12, RTCM-GABP, NeuroAED}.

Typical preprocessing for direct event streams includes noise filtering, temporal clustering, event deduplication, and density normalization. Noise filtering suppresses sensor artifacts and environmental interference:
\begin{equation}
\mathcal{E}_{\mathrm{filtered}} = \{ e_i \in \mathcal{E} \mid \Phi(e_i) = 1 \},
\end{equation}
where \(\mathcal{E}\) denotes the raw event set, \(\mathcal{E}_{\mathrm{filtered}}\) denotes the filtered event set after noise suppression, \(e_i\) denotes the \(i\)-th event, and \(\Phi(\cdot)\) is a Boolean-valued validity function that returns 1 when an event is retained and 0 otherwise. Temporal clustering then groups events according to temporal proximity in order to identify meaningful motion patterns:
\begin{equation}
\mathcal{C}_j = \bigcup_{i}\{e_i \mid t_{i+1} - t_i \leq \Delta T\},
\end{equation}
where $\Delta T$ is a predefined temporal threshold. Event deduplication further merges redundant or duplicate events:
\begin{equation}
\mathcal{E}_{\mathrm{dedup}} = \operatorname{merge}(\{e_i, e_{i+1} \mid e_i \approx e_{i+1}\}),
\end{equation}
while density normalization alleviates local response imbalance across the sensor plane:
\begin{equation}
\mathcal{E}_{\mathrm{norm}} = \frac{\mathcal{E}_{\mathrm{filtered}} - \min(\mathcal{E}_{\mathrm{filtered}})}{\max(\mathcal{E}_{\mathrm{filtered}}) - \min(\mathcal{E}_{\mathrm{filtered}})}.
\end{equation}
Here, \(e_i \approx e_{i+1}\) indicates that two neighboring events are regarded as redundant according to predefined spatial, temporal, and polarity consistency criteria; \(\operatorname{merge}(\cdot)\) denotes the corresponding event-merging operation; and \(\min(\cdot)\) and \(\max(\cdot)\) are applied to the selected event-derived representation for density normalization.

Through the above operations, the direct event stream is transformed into a cleaner and more informative signal representation, thereby reducing sparsity-induced instability and enhancing the delineation of dynamic patterns relevant to pedestrian detection.

After preprocessing, the event stream can be embedded into deep learning architectures to construct a multi-dimensional representation of the scene. Such representations may include temporal event distributions, localized spatio-temporal volumes, polarity statistics, spatial depth cues, and information entropy, all of which jointly characterize pedestrian motion and scene dynamics. Formally, an example feature vector at time $t$, derived from the normalized event stream $\mathcal{E}_{\mathrm{norm}}$, can be expressed as:

\begin{equation}
\mathbf{F}_t = \operatorname{encode}(\mathcal{E}_{\mathrm{norm}}(t)).
\end{equation}

This high-dimensional feature representation enables downstream models to capture complex pedestrian behaviors and transient motion patterns with high precision.

Unlike event-to-frame conversion or frame-event fusion, direct event stream processing preserves the original temporal granularity of asynchronous event data. This is particularly beneficial for modeling fast motion, rapid luminance changes, and short-lived visual structures. Spiking Neural Networks (SNNs) and related event-native architectures are especially suitable in this setting, as they can directly exploit event timing information for motion-sensitive detection. In practice, this processing route typically involves suppressing spurious events, stabilizing event density across the visual field, and encoding the refined stream into a representation that preserves temporal granularity while remaining suitable for neural inference.

The primary advantage of direct event stream processing lies in its ability to handle raw, high-frequency signals without collapsing them into frame-like structures. This makes it particularly suitable for capturing subtle and transient pedestrian dynamics under high-speed motion and challenging illumination, where conventional frame-based approaches often suffer from motion blur, temporal aliasing, or data redundancy.

\begin{figure}[t]
    \centering
    \includegraphics[width=0.7\linewidth]{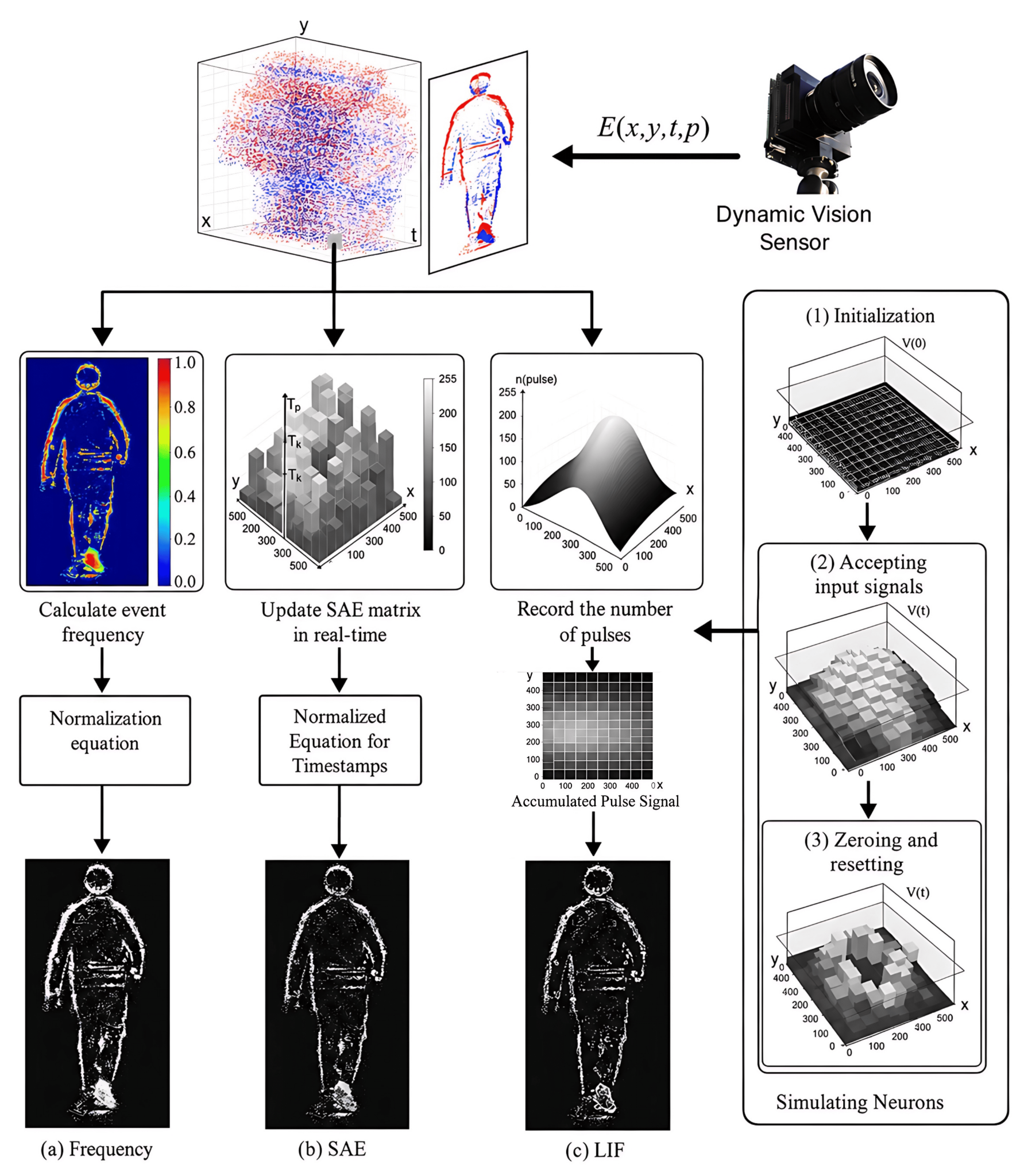}
    \caption{The conversion of event streams into event frames can be understood as projecting asynchronous event signals within a temporal window onto a frame-like representation: (a) assigning pixel luminance according to event frequency, (b) encoding the temporal order of events according to their occurrence within a selected time interval, and (c) converting accumulated events into image intensity through a neuron-inspired integration mechanism.}
    \label{333}
\end{figure}

\subsubsection{Event-to-Frame Conversion}

Event-to-frame conversion transforms asynchronous event data into structured frame-like representations that can be directly processed by conventional computer vision models. This strategy preserves part of the temporal dynamics of event data while improving compatibility with mature frame-based detection architectures. At present, this remains the most widely adopted processing paradigm in EB-PD.

The predominant event-to-frame encoding strategies can be broadly categorized into three types \cite{EventB2021}, as illustrated in Fig.~\ref{333}: frequency-based event encoding \cite{10}, active-event-surface-based encoding \cite{14,15}, and Leaky Integrate-and-Fire (LIF)-based event accumulation \cite{16,17,18}. Despite their differences, these methods share a common objective: to convert sparse asynchronous events into structured representations that are amenable to conventional feature extraction pipelines.

\begin{itemize}
  \item $t_i$ denotes the timestamp of each event, which determines its temporal contribution to the generated frame;
  \item $(x_i, y_i)$ are the spatial coordinates corresponding to the activated pixel location;
  \item $p_i \in \{-1, +1\}$ denotes the polarity of the intensity variation.
\end{itemize}

The event-to-frame conversion process usually includes four closely related operations: event integration, polarity fusion, spatial coherence enhancement, and inter-frame continuity modeling. In the following formulations, \(I_t(x,y)\), \(P_t(x,y)\), \(S_t(x,y)\), and \(F_t(x,y)\) denote the temporally integrated event image, polarity-fused map, spatially smoothed representation, and temporally stabilized event frame at time \(t\), respectively. Event integration accumulates asynchronous events within a temporal window to construct a frame-like signal:
\begin{equation}
I_t(x, y) = \sum_{e_i \in \mathcal{W}_t(x, y)} p_i \cdot \exp\left(-\frac{t - t_i}{\tau}\right),
\end{equation}
where $\mathcal{W}_t(x, y)$ denotes the set of events at pixel $(x,y)$ around time $t$, and $\tau > 0$ is the temporal decay constant controlling the attenuation of past events. On this basis, polarity fusion aggregates positive and negative events to emphasize meaningful motion-induced changes while suppressing noise:
\begin{equation}
P_t(x, y) =
\begin{cases}
1, & \text{if } \sum_{e_i \in \mathcal{W}_t(x, y)} p_i > \theta_p, \\
-1, & \text{if } \sum_{e_i \in \mathcal{W}_t(x, y)} p_i < -\theta_p, \\
0, & \text{otherwise},
\end{cases}
\end{equation}
where $\theta_p$ is a predefined polarity threshold. Spatial coherence enhancement is then applied to improve local consistency while preserving structural boundaries:
\begin{equation}
S_t(x, y) = (G_{\sigma} * I_t)(x, y),
\end{equation}
where $G_{\sigma}$ denotes a Gaussian smoothing kernel with standard deviation $\sigma > 0$. Finally, inter-frame continuity is maintained through temporal filtering to improve frame-to-frame stability and reduce flickering artifacts:
\begin{equation}
F_t(x, y) = \alpha \cdot S_t(x, y) + (1 - \alpha) \cdot F_{t-1}(x, y),
\end{equation}
where $\alpha \in [0,1]$ controls the blending ratio between the current and previous frames.

While the above formulations summarize the canonical event-to-frame conversion pipeline, they are still predominantly based on handcrafted aggregation rules. Such strategies are computationally convenient and relatively easy to integrate into existing CNN-based detectors, but they often rely on fixed temporal windows and simple accumulation mechanisms, which may suppress fine-grained temporal order, distort local event statistics, and entangle motion-related and appearance-related information. To address these limitations, recent research has increasingly shifted toward more statistically grounded and task-adaptive event representations \cite{TORE2023, MAD2025, EvRepSL2024, SSER2025, EVA2025}.

In particular, TORE (Time-Ordered Recent Event) volumes explicitly preserve the recency ordering of recent events, instead of merely aggregating them within a fixed window, thereby retaining richer temporal structure for fast motion, motion boundaries, and transient object contours \cite{TORE2023}. MAD (Motion and Appearance Decoupling Representation) further argues that conventional event tensors often entangle motion saliency and appearance context, and therefore proposes a decoupled representation that separates motion and appearance components before feature interaction and fusion \cite{MAD2025}. In addition, data-driven representation learning has recently become an important direction. EvRepSL constructs event representations from spatio-temporal statistics and refines them via self-supervised learning, reducing dependence on manually designed aggregation heuristics \cite{EvRepSL2024}. Similarly, recent recurrent self-supervised representations such as SSER aim to avoid hard temporal discretization by learning asynchronous encoders directly from event timestamps and polarities, while also exhibiting strong potential for low-latency hardware deployment \cite{SSER2025}. Furthermore, EVA extends this line of research toward event-by-event asynchronous representation learning by combining asynchronous encoding with linear-attention-based self-supervised learning, suggesting a possible transition from synchronous tensorization to highly expressive asynchronous representations \cite{EVA2025}. Collectively, these studies indicate that event representation is evolving from fixed accumulation toward order-preserving, decoupled, and learnable statistical encoding, which is particularly relevant for pedestrian detection under high-speed motion, cluttered backgrounds, and adverse illumination.

In addition to the three conventional categories, Wan et al. \cite{EventB2021} introduced an innovative event representation termed Neighborhood Suppression Time Surface (NSTS). This method is inspired by time-surface generation and HATS-based processing \cite{wan1,wan2}. Its key novelty lies in suppressing local intensity responses around each pixel independently, thereby reducing the dominance of dense-event regions and improving the relative significance of sparse but informative event regions.

Moreover, \cite{Zubic2024StateSM} employed the Sparse Surface Model (SSM) to convert event streams into structured event-frame tensors and validated its effectiveness on the Gen1 \cite{Prophesee2022} and 1Mpx \cite{1-M} datasets. SSM accumulates events over fixed temporal intervals to construct dense tensor maps $H_k$, allowing convolutional layers to extract spatial features from structured event data. To preserve temporal continuity, Convolutional Long Short-Term Memory (ConvLSTM) layers are introduced, with the hidden state updated as:
\begin{equation}
h_k = F(H_k, h_{k-1}),
\end{equation}
where $F$ denotes the ConvLSTM operation.

For detection, multi-scale feature extraction layers are used to predict bounding boxes $B_k$ and $B'_{k+1}$, which are aligned with the corresponding ground-truth boxes $B^*_k$ and $B^*_{k+1}$. The overall training objective can be formulated as:
\begin{equation}
\mathcal{L} = \mathcal{L}_{\mathrm{cls}} + \mathcal{L}_{\mathrm{reg}} + \mathcal{L}_{\mathrm{aux}},
\end{equation}
where \(\mathcal{L}\) denotes the total training loss, \(\mathcal{L}_{\mathrm{cls}}\) denotes the classification loss, \(\mathcal{L}_{\mathrm{reg}}\) denotes the bounding-box regression loss, and \(\mathcal{L}_{\mathrm{aux}}\) denotes the auxiliary supervision loss.

SSM leverages the high temporal resolution of event cameras to capture rapid dynamics while reducing redundancy in data representation. In parallel, Spiking Neural Networks (SNNs) remain an important family of event-native models, as they mimic the biological spiking mechanism and can process events at fine temporal granularity. The output spike train of a neuron can be expressed as:
\begin{equation}
y(t) = \sum_{i=1}^{N} w_i \cdot s(t - t_i),
\end{equation}
where $y(t)$ denotes the continuous output response, $w_i$ are synaptic weights, $t_i$ are spike times, and $s(\cdot)$ is the spike response function, typically modeled as a causal decaying kernel.

Overall, event-to-frame conversion establishes an effective bridge between the high temporal resolution of event streams and the rich ecosystem of mature frame-based vision architectures \cite{EF1,EF2,EF3}. Although some temporal precision is inevitably sacrificed during conversion, this paradigm remains highly practical for EB-PD owing to its strong compatibility with advanced detection backbones and efficient deployment frameworks \cite{EF4}.

\subsubsection{Joint Event-Frame Fusion}

Joint event-frame fusion seeks to combine the high temporal sensitivity of event data with the rich spatial context provided by conventional image frames \cite{PEF1,PEF2,PEF3}. This multimodal strategy is particularly attractive in pedestrian detection, as it enables complementary modeling of motion dynamics and appearance information.

Recent multimodal representation learning has moved beyond conventional synchronized two-branch fusion. In particular, ACGR (Asynchronous Collaborative Graph Representation) proposes a unified graph-based paradigm for jointly representing frames and events, rather than first converting event data into image-like tensors tied to the frame rate \cite{ACGR2025}. By constructing sparse unimodal graphs for both modalities and aligning them through asynchronous collaborative modules, ACGR preserves the spatio-temporal sparsity of event data while alleviating inter-modal misalignment. This insight is especially relevant for EB-PD, as it suggests that parallel event-frame processing need not be restricted to frame-rate-limited fusion, but can instead evolve toward unified asynchronous representations with both high efficiency and strong detection potential.

\begin{algorithm}
\caption{Fusion Algorithm for Event and Frame Stream Data}
\begin{algorithmic}[1]
\Require{$E$: Stream of event tuples $(t, x, y, p)$, $F$: Sequence of frames, $\tau$: Synchronization threshold}
\Ensure{$D_{\mathrm{fused}}$: Fused data for model ingestion}

\Procedure{FuseStreams}{$E$, $F$, $\tau$}
    \State $E_{\mathrm{pre}}\gets \Call{PreprocessEvents}{E}$
    \State $F_{\mathrm{pre}}\gets \Call{PreprocessFrames}{F}$
    \State $E_{\mathrm{sync}}, F_{\mathrm{sync}}\gets \Call{Synchronize}{E_{\mathrm{pre}}, F_{\mathrm{pre}}, \tau}$
    \State Initialize $D_{\mathrm{fused}}$ to an empty list
    \For{$t = 1$ to $T$}
        \State $\mathrm{Feat}_{E}\gets \Call{ExtractEventFeatures}{E_{\mathrm{sync}}[t]}$
        \State $\mathrm{Feat}_{F}\gets \Call{ExtractFrameFeatures}{F_{\mathrm{sync}}[t]}$
        \State $\mathrm{Feat}_{\mathrm{fused}}\gets \Call{FuseFeatures}{\mathrm{Feat}_{E}, \mathrm{Feat}_{F}}$
        \State Append $\Call{Standardize}{\mathrm{Feat}_{\mathrm{fused}}}$ to $D_{\mathrm{fused}}$
    \EndFor
    \State \Return $D_{\mathrm{fused}}$
\EndProcedure
\end{algorithmic}
\label{algorithmic1}
\end{algorithm}

Algorithm~\ref{algorithmic1} \cite{SSDE-RDVS,MFE-FPD,PEF2} presents a representative example of simultaneous frame-event input, which has become a cornerstone of advanced EB-PD systems. The key idea is to harmonize frame-based spatial cues with event-based temporal cues so as to produce a richer and more robust data representation for pedestrian detection.

The fusion process begins with the independent preprocessing of event and frame streams. For the event stream, preprocessing focuses on suppressing noise and removing irrelevant signals:
\begin{equation}
E_{\mathrm{filtered}} =
\{e \in E \mid \operatorname{NoiseFilter}(e) \land \operatorname{SignalThreshold}(e)\},
\end{equation}
where \(E\) denotes the raw event stream, \(E_{\mathrm{filtered}}\) denotes the filtered event set after noise and irrelevant-signal suppression, and \(\operatorname{NoiseFilter}(\cdot)\) and \(\operatorname{SignalThreshold}(\cdot)\) denote the event validity tests used in preprocessing. In parallel, frame data undergo spatial enhancement to improve texture clarity and structural detail.

The central challenge is temporal synchronization, since event streams are asynchronous while frame streams are sampled at fixed intervals. A temporal alignment operation may be expressed as:
\begin{equation}
\{E_{\mathrm{sync}}, F_{\mathrm{sync}}\} =
\operatorname{TemporalAlign}(E_{\mathrm{filtered}}, F, \tau),
\end{equation}
where \(F\) denotes the frame sequence, \(E_{\mathrm{sync}}\) and \(F_{\mathrm{sync}}\) denote the temporally synchronized event and frame streams, respectively, and \(\tau\) denotes the synchronization tolerance window.

Once aligned, feature extraction can exploit both the dynamic information captured by events and the contextual appearance information available in frames:
\begin{equation}
\mathrm{Features} =
\operatorname{ExtractFeatures}(E_{\mathrm{sync}}, F_{\mathrm{sync}}),
\end{equation}
thereby constructing a unified spatio-temporal feature representation.

This integration strategy highlights the methodological principle underlying parallel event-frame processing: events contribute fine temporal sensitivity, while frames provide dense spatial context. Their complementary combination enables a richer and more discriminative representation of pedestrian behavior in complex and dynamic environments.

\subsection{Preprocessing}

The preprocessing stage in the EB-PD pipeline aims to improve data integrity and representation quality for each processing paradigm, thereby facilitating more effective feature extraction and downstream recognition. More generally, in event-based vision, preprocessing should be regarded not merely as a task-specific auxiliary step for pedestrian detection, but as a fundamental upstream component of event-stream perception. Its primary goals include suppressing spurious events and background activity noise, preserving informative spatio-temporal correlations, stabilizing event density across space and time, and improving robustness under heterogeneous sensing conditions.

From a broader methodological perspective, generic event-stream preprocessing and denoising methods can be organized into four representative categories: \emph{statistical/probabilistic denoising}, \emph{noise-specific lightweight filtering}, \emph{learning-based spatio-temporal denoising}, and \emph{task- or scenario-aware extensions}. As summarized in Table~\ref{tab:preprocess_taxonomy}, these categories exhibit different trade-offs in interpretability, adaptability, computational cost, and deployment suitability. In parallel, recent resources such as LED and E-MLB have enabled more systematic training and evaluation of event denoising methods across diverse scenes and noise levels \cite{Duan2024LED,Ding2024EMLB}.

\begin{table}[t]
\centering
\caption{Principle-based comparison of generic event-stream denoising methods relevant to EB-PD.}
\label{tab:preprocess_taxonomy}
\scriptsize
\setlength{\tabcolsep}{3pt}
\renewcommand{\arraystretch}{1.02}
\begin{tabularx}{\linewidth}{
>{\raggedright\arraybackslash}p{2.8cm}
>{\raggedright\arraybackslash}p{3.4cm}
>{\raggedright\arraybackslash}p{3.30cm}
>{\raggedright\arraybackslash}p{4.00cm}
}
\toprule
\textbf{Category} & \textbf{Representative methods} & \textbf{Advantages} & \textbf{Limitations} \\
\midrule

\makecell[l]{Statistical denoising}
& \makecell[l]{PUGM \cite{Wu2021PUGM}}
& \makecell[l]{Interpretable;\\training-free;\\preserves local consistency}
& \makecell[l]{Rigid assumptions;\\weaker adaptability to\\complex noise} \\
\addlinespace[4pt]

\makecell[l]{Lightweight filtering}
& \makecell[l]{BA denoising \cite{Guo2023BALowCost}}
& \makecell[l]{Low cost; \\low latency;\\hardware-friendly}
& \makecell[l]{Mainly targets BA noise;\\limited generality} \\
\addlinespace[4pt]

\makecell[l]{Learning-based denoising}
& \makecell[l]{GNN-Transformer \cite{Alkendi2024GNNTransformerDenoise};\\
EDformer \cite{Jiang2024EDformer};\\
EDmamba \cite{Ruan2025EDmamba};\\
ASTEDNet \cite{Wu2024ASTEDNet};\\
DBRGNN \cite{Feng2025DBRGNN};\\
Point-cloud noise modeling \cite{Annamalai2024BeyondSupervision}}
& \makecell[l]{Adaptive; \\robust to varied noise; \\captures complex dependencies}
& \makecell[l]{Higher computation;\\data-dependent;\\some methods need\\broader validation} \\
\addlinespace[4pt]

\makecell[l]{Task-aware extensions}
& \makecell[l]{SMNE \cite{Shiba2025SMNE};\\
AWTS \cite{Yin2025AWTS};\\
Controlled noise injection \cite{Kowalczyk2025LearningFromNoise};\\
SPIE denoising method \cite{Lv2024EventCameraDenoising}}
& \makecell[l]{Better task alignment;\\useful in specialized settings}
& \makecell[l]{Narrower generality;\\higher modeling complexity} \\

\bottomrule
\end{tabularx}
\end{table}

For direct event-stream processing, preprocessing is primarily centered on temporal fidelity. Adaptive noise filtering purifies the event stream \cite{E-PD-DVS}, while event deduplication and event aggregation strategies expose meaningful temporal structures \cite{Near-Chip}. Event density normalization further improves the consistency of temporal feature extraction \cite{zao1}. In this paradigm, preprocessing must operate directly on sparse asynchronous events, which makes the pipeline particularly sensitive to isolated spurious activations and background activity noise. Statistical and probabilistic methods are advantageous when interpretability and local event-structure preservation are emphasized, whereas lightweight background-activity suppression is preferable when low latency and deployment efficiency are the dominant constraints \cite{Wu2021PUGM,Guo2023BALowCost}. More recent direct-stream denoisers, including asynchronous spatio-temporal neural denoising, residual graph neural networks, and unsupervised spatio-temporal point-cloud noise modeling, further show that denoising can be made more adaptive while remaining closer to the native asynchronous structure of event streams \cite{Wu2024ASTEDNet,Feng2025DBRGNN,Annamalai2024BeyondSupervision}.

For event-to-frame conversion, preprocessing transforms asynchronous events into structured frame-like signals \cite{zao4}, thereby ensuring temporal continuity at the representation level. Through event accumulation and polarity encoding, motion saliency is enhanced. Spatial filtering \cite{zao1,zao2,zao3} and related enhancement operations are subsequently applied to improve frame quality and highlight pedestrian-related structures. Regularization also plays an important role in ensuring compatibility with mature image-based processing pipelines \cite{zhengze1,zhengze2}. Here, however, the quality of the generated representation depends critically on whether noisy events are sufficiently suppressed before accumulation. In this sense, event-to-frame preprocessing is no longer limited to fixed-window accumulation and heuristic denoising, but is gradually evolving toward noise-adaptive and statistically informed representation refinement. Learning-based denoisers such as EDformer are particularly attractive in this setting because they improve robustness under heterogeneous noise conditions, while more efficient state-space formulations such as EDmamba are promising when computational efficiency must also be considered \cite{Jiang2024EDformer,Ruan2025EDmamba}. Scenario-specific strategies, such as attribute-weighted time-surface denoising, further indicate that local event reliability estimation remains useful when refining event frames under challenging sensing conditions \cite{Yin2025AWTS}.

For joint event-frame fusion, preprocessing aims to achieve a synergistic integration of temporally dense event streams and spatially informative image frames \cite{SSDE-RDVS}. This requires precise temporal synchronization to align the two modalities \cite{pal1,pal2}. Feature standardization is then employed to ensure a consistent cross-modal representation \cite{MFE-FPD}. Such careful alignment is essential for the extraction of integrated spatio-temporal features \cite{pal3}, which are critical for capturing pedestrian motion under complex real-world conditions. Compared with single-branch pipelines, parallel event-frame systems can partially tolerate moderate event noise because the frame stream provides complementary appearance cues. Nevertheless, they are also vulnerable to misalignment errors introduced by unstable event preprocessing. Accordingly, in this setting preprocessing should emphasize not only synchronization, but also stable denoising and representation consistency prior to fusion. In addition, recent studies suggest that preprocessing may be further strengthened by coupling denoising with motion estimation or robustness-oriented noise modeling, rather than treating denoising as a fully isolated step \cite{Shiba2025SMNE,Kowalczyk2025LearningFromNoise}.

\subsection{Feature Extraction and Representation}

Feature engineering is a key stage in the EB-PD pipeline, in which preprocessed signals are transformed into discriminative and informative descriptors suitable for pedestrian detection \cite{tezheng3,tezheng1,tezheng2}.

For direct event-stream processing, feature extraction primarily leverages the temporal distribution of events to characterize dynamic scene variations. Localized spatio-temporal volumes capture micro-movements in confined regions \cite{NeuroAED}, thereby revealing subtle pedestrian behaviors. Polarity statistics are useful for modeling edges and motion direction \cite{tezheng1,RTCM-GABP}, while spatial depth-related cues may further improve scene understanding in three-dimensional space.

For event-to-frame conversion, feature extraction benefits from the temporal continuity embedded in the generated frame sequence. Motion formation cues \cite{PoT-QY-PD,GMVDT-NVS}, derived from polarity changes and accumulated event responses, reveal pedestrian trajectories and object motion patterns. The transformation process also recovers spatial textures and structural contours \cite{tezheng4}, enabling conventional detectors to model pedestrian appearance more effectively. Multi-scale analysis further improves robustness by capturing fine and coarse features simultaneously. More recent representation designs improve feature quality at the source: order-preserving encodings retain recency cues valuable for motion boundaries and rapid limb movement, decoupled representations separate motion saliency from appearance context, statistical/self-supervised encoders provide more stable descriptors under background clutter and illumination changes, and asynchronous representations reduce the loss of high-frequency temporal structure caused by forced discretization or frame-rate synchronization \cite{TORE2023,MAD2025,EvRepSL2024,SSER2025,EVA2025,ACGR2025}.

For joint event-frame fusion, feature extraction and representation aim to integrate detailed temporal descriptors from event streams with dense spatial descriptors from image frames \cite{SSDE-RDVS}. Spatio-temporal fusion techniques combine motion and appearance into a unified representation. By jointly modeling polarity, texture, and contextual cues from both modalities \cite{MFE-FPD}, such methods build more comprehensive feature descriptors for pedestrian perception. In addition, features for dynamic background adaptation help maintain model sensitivity under challenging environmental variations.

\subsection{Detection Models}

The final stage of the EB-PD pipeline concerns the design of detection models capable of exploiting the feature representations generated in the preceding stages. It should be emphasized that the effectiveness of these networks depends not only on the backbone architecture itself, but also on the quality of the underlying event representation, particularly whether temporal order, local statistics, and motion cues are adequately preserved during encoding \cite{TORE2023,MAD2025,EvRepSL2024}.

CNNs have been widely adopted in EB-PD tasks, with representative architectures including YOLOv7 \cite{HR-EC-PD}, YOLOv5 \cite{E-YOLO-FPS,PoT-QY-PD}, YOLOv3 \cite{E-PD-DVS,MFE-FPD,Mixed2019}, and YOLOv3-Tiny \cite{MC-EIF-PD,Mixed2019}. These architectures are attractive because they provide a strong trade-off between detection accuracy, inference speed, and deployment efficiency. For example, YOLOv7 \cite{yolov7} and YOLOv5 \cite{yolov5} are known for high accuracy and real-time performance, making them suitable for latency-sensitive pedestrian detection scenarios. YOLOv3 \cite{yolov3} and YOLOv3-Tiny \cite{yolotiny}, while earlier architectures, still provide an effective balance between detection performance and computational cost, especially on resource-constrained platforms.

Algorithm~\ref{algorithmic1211} outlines the general detection logic of the YOLO family when adapted to EB-PD. In such models, the input image is divided into an $S \times S$ grid, and each grid cell predicts $B$ bounding boxes together with class probabilities. This grid-based formulation is well suited to event-driven detection systems that require efficient inference and rapid localization.

In addition to mainstream CNN detectors, alternative models have also been explored to address the unique characteristics of EB-PD. For instance, the Genetic Algorithm-Back Propagation (GA-BP) neural network \cite{RTCM-GABP} combines the global optimization ability of genetic algorithms with the learning efficiency of backpropagation, thereby improving robustness in complex environments. Likewise, the Spatial Attention Model (SAM) \cite{EOD-LSAM} improves detection performance by emphasizing informative spatial features through attention-based reweighting.

\begin{algorithm}[H]
\caption{Advanced YOLO Algorithm for EB-PD}
\begin{algorithmic}[1]
    \State \textbf{Input:} Image $I$ divided into $S \times S$ grid, Event stream $E$
    \State \textbf{Output:} Refined bounding boxes, confidence scores, class probabilities
    \Procedure{YOLO Detection}{}
        \For{each cell $(i,j)$ in $S \times S$ grid}
            \For{each bounding box $b$ in cell}
                \State Predict offsets $\Delta x, \Delta y, \Delta w, \Delta h$
                \State Calculate box center coordinates:
                \State $x = \sigma(\Delta x) + i$
                \State $y = \sigma(\Delta y) + j$
                \State Calculate width and height:
                \State $w = p_w e^{\Delta w}$
                \State $h = p_h e^{\Delta h}$
                \State Compute confidence score:
                \State $C_b = \sigma(\text{Output for IoU})$
                \State Predict class probabilities $\{p_1, p_2, \dots, p_C\}$
                \State $p_k = \sigma(\text{class score output for } k)$
                \State \textbf{Integrate Event Data:}
                \State Update $(x, y, w, h)$ using events in $E$ near $(i,j)$
                \State \textit{Event-driven adjustments:}
                \State $t_x', t_y' = \text{AdjustCoordinates}(E, x, y, \Delta t)$
                \State $w', h' = \text{AdjustDimensions}(E, w, h, \Delta t)$
                \State \textit{Noise filtering and temporal consistency:}
                \State Apply Kalman filtering to $(x, y, w, h)$ over time
                \State Use historical data to refine confidence $C_b$
            \EndFor
            \State [Apply non-maximum suppression]
            \State [Use event-based spatial attention]
        \EndFor
    \EndProcedure
\end{algorithmic}
\label{algorithmic1211}
\end{algorithm}

\section{PREDICTION MODELS}

To facilitate a structured comparison, Table~\ref{method} summarizes representative EB-PD methods in terms of processing paradigm, model family, time window, dataset, and reported performance. However, these methods should not be interpreted merely as isolated implementations. Rather, they occupy different operating points on a shared trade-off surface spanning temporal fidelity, detection accuracy, computational efficiency, and deployment complexity. A critical review of EB-PD therefore requires not only method listing, but also cross-paradigm comparison.

To clarify the organization of this section, we first provide a cross-paradigm comparison of representative EB-PD methods from the perspectives of temporal fidelity, detection accuracy, computational efficiency, and deployment complexity. This comparison establishes the general trade-off principles among direct event-stream processing, event-to-frame conversion, and joint event-frame fusion. The subsequent subsections then apply these principles to two major application settings: dynamic traffic scenes and static surveillance scenes. In this way, the cross-paradigm analysis serves as the conceptual basis for the following scene-oriented discussions rather than as an isolated subsection.

\begin{table}[H]
\centering
\caption{Summary of 19 representative EB-PD methods grouped by processing paradigm. ``ES'' denotes direct event-stream processing, ``ES to EF'' denotes event-to-frame conversion, ``ES+F'' denotes joint event-frame fusion, and ``F to ES'' denotes frame-to-event conversion.}
\label{method}
\rotatebox{90}{%
\begin{minipage}{\textheight}
\renewcommand{\arraystretch}{1.2}
\footnotesize
\setlength{\tabcolsep}{5pt}
\begin{tabular}{
|>{\centering\arraybackslash}p{2.6cm}
|>{\centering\arraybackslash}p{1.8cm}
|>{\centering\arraybackslash}p{1.5cm}
|>{\centering\arraybackslash}p{2.3cm}
|>{\centering\arraybackslash}p{2.5cm}
|>{\centering\arraybackslash}p{1.8cm}
|>{\centering\arraybackslash}p{2.2cm}
|>{\centering\arraybackslash}p{2.5cm}|
}
\hline
\textbf{Methods} & \textbf{Publication} & \textbf{Family} & \textbf{Network} & \textbf{Metric \& Performance} & \textbf{Time Window} & \textbf{Processing Paradigm} & \textbf{Dataset} \\
\hline \hline
HR-EC-PD \cite{HR-EC-PD} & PP-RAI 2023 & CNN & YOLOv7 & 67.7\% mAP@0.5 IoU & — & ES to EF & 1Mpx \\
PoT-QY-PD \cite{PoT-QY-PD} & DSD-E 2023 & CNN & YOLOv5 & 74.3\% mAP@0.5 IoU & 10 ms & ES to EF & 1Mpx \\
GMVDT-NVS \cite{GMVDT-NVS} & IoT-J 2019 & CNN & YOLOv3 & — & — & ES to EF & Real-world test \\
MC-EIF-PD \cite{MC-EIF-PD} & Neurorobot 2019 & Hybrid & YOLOv3 \& YOLOv3-Tiny & 82.28\% mAP@0.5 IoU & 20 ms & ES to EF & Self-collected \\
PDTS-DVS \cite{PDTS-DVS} & CyS 2018 & Hybrid & DBscan & — & 40 ms & ES to EF & Self-collected \\
E-PD-DVS \cite{E-PD-DVS} & Electronics 2021 & CNN & YOLOv3 & 87.43\% mAP & 10 ms & ES to EF & Self-collected \\
DVS-RGB-Fusion \cite{DVS-RGB-Fusion} & VISAPP 2025 & CNN & YOLOv4 & 91.6\% mAP & 8 ms & ES to EF & Self-collected \\
EB-FZ-PD \cite{EB-FZ-PD} & Vehicles 2025 & CNN & YOLOv8 & 83.0\% mAP@0.5 IoU & 2.4–33.4 ms & ES to EF & Gen1 \\
DS-FL-PD \cite{DS-FL-PD} & TIV 2024 & CNN & DTSDNet & 49.3\% mAP@0.5 IoU & 10–50 ms & ES to EF & Gen1 \& 1Mpx \\
\hline \hline
RTCM-GABP \cite{RTCM-GABP} & AUTEEE 2021 & Hybrid & GA-BP & 70.8\% mAP@0.5 IoU & — & ES & Self-collected \\
NeuroAED \cite{NeuroAED} & TIFS 2020 & CNN & EMST & 93.3\% mAP & — & ES & Self-collected \\
EOD-LSAM \cite{EOD-LSAM} & ICARM 2021 & Hybrid & SAM & 35.5\% mAP & — & ES & Gen1 \& 1Mpx \\
HR-BZ-PD \cite{HR-BZ-PD} & CVPRW 2025 & SNN & anchor-free detection & 33.6\% mAP@0.3 IoU & 7.1 ms & ES & PEDRo \\
NSL-CA-PD \cite{NSL-CA-PD} & NCE 2025 & SNN & FOMO-MobileNetV2 & — & 8.21–24.62 ms & ES & Self-collected \\
SNN-SK-PD \cite{SNN-SK-PD} & Electronics 2024 & SNN & SPS ResNet18 & — & — & ES & Self-collected \\
\hline \hline
SSDE-RDVS \cite{SSDE-RDVS} & ACPR 2021 & Hybrid & CNN & — & — & ES+F & Virtual \\
MFE-FPD \cite{MFE-FPD} & ICRA 2019 & CNN & YOLOv3 \& YOLOv3-Tiny & 93\% mAP & — & ES+F & Self-collected \\
DAGr-PD \cite{DAGr-PD} & Nature 2024 & DAGr & YOLOX & 41.9\% mAP@0.5 IoU & — & ES+F & DSEC-Detection \\
\hline \hline
E-YOLO-FPS \cite{E-YOLO-FPS} & IoT-J 2023 & CNN & YOLOv5 & 69\% mAP@0.2 IoU & — & F to ES & Virtual \\
\hline
\end{tabular}
\end{minipage}
}
\end{table}

\subsection{Cross-Paradigm Comparison and Trade-off Analysis}

A first major trade-off concerns the representation pathway itself. Direct event-stream methods preserve the native asynchronous nature of EVS and therefore best retain temporal fidelity, motion sharpness, and low-latency responsiveness. This makes them conceptually well aligned with the sensing principle of EVS, particularly in high-speed or rapidly changing scenes. However, raw event streams are sparse, irregular, and difficult to process with conventional vision backbones, which often leads to weaker architectural maturity and less stable detection performance in practice. By contrast, event-to-frame methods sacrifice part of the asynchronous structure in exchange for compatibility with mature frame-based detectors. Their primary advantage is practical accuracy and engineering accessibility: once events are converted into frame-like representations, a large ecosystem of CNN detectors can be reused effectively. The cost of this convenience is that temporal sparsity, microsecond timing, and part of the latency advantage of EVS are weakened during aggregation. Event-frame fusion methods occupy an intermediate position. They exploit the temporal sensitivity of events together with the dense spatial context of conventional frames, and thus often provide stronger robustness under nighttime, low-light, or cluttered conditions. However, this gain comes at the expense of higher system complexity, synchronization overhead, and potentially heavier deployment requirements.

A second and more fundamental trade-off arises between SNN-based and CNN-based processing. SNNs are, in principle, more faithful to event cameras because they operate naturally on sparse asynchronous spikes, avoid forced discretization, and are better aligned with neuromorphic hardware. In theory, this makes them appealing for low-power, event-native, and latency-sensitive perception. Yet the current EB-PD literature also suggests clear limitations: training remains more difficult, large-scale benchmark evidence is still limited, and practical accuracy often trails strong CNN-based event-frame pipelines. In contrast, CNNs currently benefit from substantially greater architectural maturity, stronger optimization stability, and a richer ecosystem of reusable detection backbones. This explains why many of the best-performing EB-PD methods still rely on event-to-frame conversion or event-frame fusion followed by CNN-style detectors. Therefore, the current state of the field does not support the simplistic conclusion that SNNs are universally superior. Rather, SNNs are more faithful to the sensing principle of EVS, whereas CNN-based pipelines currently benefit from stronger practical accuracy and engineering maturity.

A third critical trade-off, especially for event-to-frame approaches, concerns the temporal accumulation window $\Delta T$. Increasing $\Delta T$ generally improves event density, signal-to-noise ratio, and representation stability, which is beneficial for frame-based detectors. However, a larger window also increases effective delay, compresses temporal ordering, and may introduce motion smearing or temporal aliasing in fast scenes. Conversely, reducing $\Delta T$ preserves temporal responsiveness and maintains more of the low-latency advantage of EVS, but often produces sparse and noisy frame representations that are harder to detect robustly. Therefore, event-to-frame EB-PD is fundamentally a trade-off between \emph{temporal fidelity} and \emph{detector compatibility}. A simplified interpretation is that the effective latency of an event-to-frame detector can be approximated as
\begin{equation}
L_{\mathrm{eff}} \approx \frac{\Delta T}{2} + L_{\mathrm{inf}},
\end{equation}
where \(L_{\mathrm{eff}}\) denotes the approximate effective latency of the event-to-frame detector, \(\Delta T\) denotes the temporal accumulation window, and \(L_{\mathrm{inf}}\) denotes the network inference time. Although this is not a strict universal formula, it provides an intuitive explanation of why temporal aggregation may erode the intrinsic low-latency benefit of EVS if the accumulation window is chosen too conservatively.

Taken together, existing EB-PD methods should not be viewed as merely different implementations, but rather as different operating points on a shared design surface defined by temporal fidelity, detection accuracy, computational efficiency, and deployment complexity. From this perspective, SNN-based direct-event methods are attractive when event nativeness, energy efficiency, and latency are prioritized; CNN-based event-frame methods remain appealing when accuracy, stability, and reuse of mature detectors are the main objectives; and fusion-based methods are especially valuable when robustness across difficult illumination and dynamic backgrounds is more important than architectural simplicity. These general trade-off principles also provide the basis for the following scene-oriented analysis. In dynamic traffic scenes, latency, robustness to motion, and illumination adaptability are usually dominant concerns, whereas in static surveillance scenes, stable long-duration observation, compact feature extraction, and reliable scene understanding become more important. Therefore, the following two subsections discuss how the above cross-paradigm trade-offs are reflected in dynamic and static EB-PD applications, respectively.

\subsection{Pedestrian Detection for Dynamic Scenes}

Dynamic-scene EB-PD, especially in autonomous driving, is fundamentally governed by a latency--robustness--accuracy trade-off. In such scenarios, the detector must respond quickly to abrupt scene changes, while remaining reliable under motion blur, severe illumination transitions, and environmental noise. As a result, existing methods can be interpreted less as isolated algorithmic proposals than as different strategies for balancing temporal responsiveness against representational stability.

A dominant strategy in this setting is to convert event streams into frame-like representations and then exploit mature CNN detectors. This design is attractive because it allows EB-PD to inherit the optimization maturity, architectural stability, and practical accuracy of frame-based detection pipelines. For example, \cite{EventB2021} introduces an asynchronous feature extraction framework on top of event-frame conversion, achieving approximately 26 FPS and 87.43\% AP on a real-world dataset. Likewise, \cite{E-PD-DVS} argues that direct application of raw event streams to conventional object detectors is often impractical, and therefore proposes an improved event-to-frame conversion strategy together with intermediate-feature reuse, again reaching 26 FPS and 87.43\% accuracy. These methods illustrate a recurring pattern in dynamic-scene EB-PD: sacrificing part of the native asynchrony of EVS in order to obtain stronger detector maturity and more stable engineering performance. Their practical value is evident, but so is their limitation---as the accumulation process becomes heavier, the low-latency sensing advantage of EVS is progressively weakened.

A second line of work emphasizes robustness under real-world dynamics through motion modeling or multimodal fusion. In \cite{GMVDT-NVS}, event-based noise filtering is combined with CTRV motion modeling, three-dimensional enhanced K-means clustering, and SCDEKF-based motion estimation, with validation under actual road-testing conditions. The contribution of this work lies less in detector novelty and more in showing that real-road EB-PD benefits from explicitly modeling motion uncertainty and target association. Similarly, \cite{SSDE-RDVS} integrates RGB and EVS data for semantic segmentation and depth estimation across multiple viewpoints, improving adaptability under nighttime and low-light conditions. Such methods indicate that, in dynamic scenes, robustness is often achieved not merely by strengthening the backbone, but by incorporating motion priors, multimodal complementarity, or scene-level constraints. The advantage of this paradigm is improved resilience under adverse environments; its drawback is greater system complexity, synchronization cost, and dependence on multiple sensing channels.

A third line of work seeks to preserve the event-native advantage of EVS in extreme dynamic conditions. \cite{EVSegNet2018} demonstrates that purely event-based inputs can already support semantic segmentation, especially under extreme lighting transitions and for highly dynamic targets, while \cite{PDTS-DVS} combines EVS data with human kinematic analysis to improve pedestrian detection and tracking. These studies are important because they highlight that EVS is not merely a drop-in replacement for RGB sensing, but a distinct modality whose motion selectivity can itself be algorithmically valuable. However, they also reveal a present limitation of the field: event-native methods are often more specialized, less standardized, and less supported by mature detection ecosystems than event-frame CNN pipelines.

Overall, dynamic-scene EB-PD currently favors methods that give up part of native event fidelity in exchange for stronger robustness and deployability. Event-to-frame CNN pipelines remain dominant because they provide the most mature balance of accuracy and engineering convenience, whereas event-native and fusion-based methods are most advantageous when rapid motion, illumination extremes, or adverse weather make the intrinsic sensing properties of EVS indispensable.

\subsection{Pedestrian Detection in Static Scenes}

In static-scene EB-PD, the dominant design priorities differ from those in autonomous driving. While low latency remains beneficial, the central requirement is often stable scene understanding for surveillance, target monitoring, posture-related analysis, or behavior recognition. Consequently, methods in static environments more frequently prioritize compact feature extraction, signal interpretability, and robustness to long-duration observation, rather than preserving every aspect of the event camera's microsecond temporal advantage.

One important category consists of direct event-stream analysis and lightweight decision models. For example, \cite{Near-Chip} integrates an event filtering module with a binary-neural-network-based detection module, showing that event-native preprocessing can reduce noise and simplify downstream classification. Similarly, \cite{RTCM-GABP} combines genetic optimization with backpropagation for road-target classification, while \cite{ApplicationOH} performs hierarchical clustering directly on raw event streams for object tracking. These methods reflect an efficiency-oriented design philosophy: instead of relying on deep and heavy backbones, they attempt to exploit the sparse structure of event data through compact models or direct signal-level operations. Their main strength is low computational overhead and relatively clear interpretability; their main weakness is that performance often depends strongly on handcrafted assumptions, scenario tuning, or carefully controlled acquisition conditions.

A second category relies on event-derived descriptors or shallow learned representations. For instance, \cite{FACM} extracts five target characteristics from 2D event point clouds and uses an SVM classifier for road-target recognition, reporting very high accuracy in a specific acquisition setup. Such methods demonstrate that event streams contain sufficiently rich geometric and motion cues to support discrimination even without large deep backbones. However, they also expose a recurring limitation of static-scene EB-PD: excellent results on self-collected or constrained scenarios do not necessarily imply strong generalization across broader benchmarks. In other words, descriptor-based success often reflects close alignment between feature design and acquisition setting, rather than universally transferable robustness.

A third category introduces fusion-based compensation for the incompleteness of event-only sensing. \cite{Mixed2019} observes that EVS data offer high dynamic range, low latency, and sparse motion sensitivity, but lack absolute brightness information; by combining APS frames with EVS signals through confidence-map fusion, the method improves consistency and accuracy over standard frame-based solutions. Likewise, \cite{MC-EIF-PD} employs multiple event encoding schemes together with both channel-level and decision-level fusion to enrich representation quality. The key insight of these methods is that, in static or surveillance-like settings, fusion is used less to recover raw temporal responsiveness and more to compensate for the semantic incompleteness of event-only data. The resulting gain is improved recognition robustness; the corresponding cost is increased model complexity, synchronization overhead, and dependence on carefully balanced modality interaction.

Finally, works such as \cite{NeuroAED} and \cite{AMNVS} show that static-scene EB-PD can benefit from event-specific sensing properties beyond conventional detection. \cite{NeuroAED} uses active event cuboids and EMST descriptors for efficient abnormal-event analysis, whereas \cite{AMNVS} introduces a multimode neuromorphic sensor with illumination measurement capability and reports data transmission rates far exceeding those of frame-based cameras. These studies reinforce an important point: event cameras are not only alternative detector inputs, but also sensing devices that may enable different formulations of scene understanding, especially when sparse motion saliency is more informative than dense appearance.

Overall, static-scene EB-PD tends to favor compact feature extraction, direct signal exploitation, and multimodal compensation rather than purely latency-driven design. Compared with dynamic-scene methods, the key trade-off here is less about preserving every microsecond of temporal precision and more about whether event-native sparsity can be converted into reliable scene understanding without over-relying on handcrafted assumptions or narrowly tailored acquisition setups.

\section{DATASETS AND EVALUATION METRICS}
\label{sec:evaluation_metrics_and_datasets}

While datasets specifically designed for event-based pedestrian detection remain limited, this survey includes both dedicated pedestrian-detection datasets and broader event-based benchmarks that are relevant to EB-PD. Since acquisition conditions alone are insufficient for assessing the practical usefulness of a dataset in this field, we review both the sensing settings and the annotation resources available for downstream perception tasks. We then summarize the evaluation metrics most commonly used in EB-PD.

\subsection{Datasets}

Table~\ref{tab:dataset-comparison} summarizes the acquisition settings and scenario characteristics of the datasets reviewed in this survey, including sensing modality, temporal resolution, environmental conditions, and image resolution. Table~\ref{tab:dataset-scale} further reports annotation-related statistics, including annotation targets, category labels, number of classes, annotation quantity, and overall dataset scale. It should be noted that not all surveyed datasets are standard 2D pedestrian-detection benchmarks; several remain highly relevant to EB-PD through tasks such as pose estimation, geometry-aware perception, activity recognition, and multimodal event-frame learning. Accordingly, the field ``\#BBoxes / annotations'' is used in a broad sense to cover bounding boxes, pose labels, activity labels, semantic annotations, or aligned frames when appropriate.

\begin{sidewaystable}[htbp]
\centering
\renewcommand{\arraystretch}{1.6}
\scriptsize
\caption{Comparison of dataset acquisition and scenario characteristics.}
\label{tab:dataset-comparison}
\begin{adjustbox}{max width=\textheight}
\begin{tabular}{
|c
@{\hskip 6pt}|>{\centering\arraybackslash}p{0.9cm}
@{\hskip 6pt}|>{\centering\arraybackslash}p{1.6cm}
@{\hskip 6pt}|>{\centering\arraybackslash}p{2.4cm}
@{\hskip 6pt}|>{\centering\arraybackslash}p{1.9cm}
@{\hskip 6pt}|>{\centering\arraybackslash}p{2.6cm}
@{\hskip 6pt}|>{\centering\arraybackslash}p{1.5cm}
@{\hskip 6pt}|>{\centering\arraybackslash}p{2.4cm}
@{\hskip 6pt}|>{\centering\arraybackslash}p{1.4cm}
@{\hskip 6pt}|>{\centering\arraybackslash}p{1.8cm}|
}
\hline
\textbf{Dataset} & \textbf{Year} & \textbf{Venue} & \textbf{Sensors} & \textbf{Temporal / Data Rate} & \textbf{Scene} & \textbf{Capture period} & \textbf{Weather} & \textbf{Season} & \textbf{Image Resolution} \\
\hline
PEDRo \cite{PEDRo2023} & 2023 & CVPRW 2023 & DAVIS346 & 50 kbps & woods, beaches, lakes, indoor scenarios & day \& night & sunny, rainy, snowy & all seasons & 346$\times$260 \\
GEN1 \cite{Prophesee2022} & 2020 & cs.CV 2020 & GEN1 + grayscale camera & 12 kbps + 120 fps & city, highway, countryside, villages, suburbs & day \& night & sunny, cloudy, rainy, snowy & all seasons & 304$\times$240 \\
Henri \cite{HSHDRV2022} & 2021 & PAMI 2021 & GEN3 + HUAWEI P20 Pro & 150 kbps + 240 fps & roads, tunnels, highways, mountain roads & day \& night & sunny, cloudy, rainy, snowy & --- & 640$\times$480 \\
PAFBenchmark \cite{Neuromo2019} & 2019 & Neurorobotics 2019 & DAVIS346 & 50 kbps & offices, corridors, walkways, plazas & day & sunny & --- & 346$\times$260 \\
FJUPD \cite{SpikeEvent2023} & 2023 & IEEE Access 2023 & DAVIS346 & 50 kbps & offices, corridors, walkways, plazas & day & sunny & --- & 346$\times$260 \\
DVS-OUTLAB \cite{DVSOUTLAB2021} & 2021 & CVPRW 2021 & CeleX-4 DVS & 200 Meps & amusement park & day \& night & rainy, cloudy, shadow & --- & 768$\times$640 \\
DHP19 \cite{DHP192019} & 2019 & CVPRW 2019 & DAVIS346 + Vicon & --- & enclosed $2\times2\times2$ m\textsuperscript{3} space & day & --- & --- & 346$\times$260 \\
NU-AIR \cite{NUAIRA2023} & 2021 & cs.CV 2021 & GEN3 & --- & intersections, campus centres, highways & day \& night & --- & --- & 640$\times$480 \\
1Mpx \cite{1-M} & 2020 & cs.CV 2020 & EB \cite{EB} + GoPro Hero6 & 300 Meps + 60 fps & city, highway, countryside, villages, suburbs & day & sunny & all seasons & 1280$\times$720 \\
MVSEC \cite{3dpp} & 2018 & RA-L 2018 & DAVIS346 + MT9V034 + VLP-16 & 50 kbps + 200 fps & interior and street view & day & sunny & --- & 346$\times$260 \\
eTraM \cite{eTraM2024} & 2024 & CVPR 2024 & Prophesee EVK4 HD & 1 $\mu$s & intersections, roadways, local streets & day, night, twilight & sunny, overcast, rainy & multi-month & 1280$\times$720 \\
SEVD \cite{SEVD} & 2024 & arXiv & DVS & --- & urban, highway, suburban, rural, intersections, roundabouts, underpasses & day, night, twilight & clear, cloudy, wet, soft-rainy, hard-rainy, foggy & --- & 1280$\times$960 \\
TUMTrafEvent \cite{TUMTrafEvent} & 2024 & IEEE T-IV 2024 & Imago VisionCam EB + Basler acA1920-50gc & 30 Meps + 25 fps & roadside gantry, urban intersection & day \& night & clear, sleet & --- & 640$\times$480 \\
HARDVS2.0 \cite{HARDVS} & 2024 & AAAI 2024 & DAVIS346 & 50 kbps & indoor, outdoor & day & --- & --- & 346$\times$260 \\
DSEC \cite{DSEC} & 2021 & RA-L 2021 & Prophesee Gen3.1 + FLIR RGB + LiDAR & 1 $\mu$s & urban, suburban, rural roads & day \& night & sunlight, shadows, low-light & --- & 640$\times$480 \\
\hline
\end{tabular}
\end{adjustbox}
\end{sidewaystable}

\begin{sidewaystable}[htbp]
\centering
\renewcommand{\arraystretch}{1.35}
\scriptsize
\caption{Comparison of dataset annotation statistics and scale.}
\label{tab:dataset-scale}
\begin{adjustbox}{max width=\textheight}
\begin{tabular}{
|>{\centering\arraybackslash}p{1.8cm}
@{\hskip 6pt}|>{\centering\arraybackslash}p{2.5cm}
@{\hskip 6pt}|>{\centering\arraybackslash}p{3.2cm}
@{\hskip 6pt}|>{\centering\arraybackslash}p{1.1cm}
@{\hskip 6pt}|>{\centering\arraybackslash}p{2.6cm}
@{\hskip 6pt}|>{\centering\arraybackslash}p{3.6cm}|
}
\hline
\textbf{Dataset} & \textbf{Main annotation target} & \textbf{Category labels} & \textbf{\#Classes} & \textbf{\#BBoxes / annotations} & \textbf{Total scale} \\
\hline
PEDRo \cite{PEDRo2023} & Human detection & person / pedestrian & 1 & 43,259 boxes & 119 records; average duration 18 s \\
GEN1 \cite{Prophesee2022} & Automotive object detection & pedestrian, car & 2 & 255,781 boxes (including 27,658 pedestrian annotations) & 39.32 h; 121 records \\
Henri \cite{HSHDRV2022} & Driving / reconstruction sequences & N/A & N/A & N/A & 44.7 GB \\
PAFBenchmark \cite{Neuromo2019} & Pedestrian / motion / fall analysis & pedestrian, action, fall events & 3 tasks & 4,670 annotated frames (pedestrian subset) & 642 records \\
FJUPD \cite{SpikeEvent2023} & Pedestrian detection & pedestrian & 1 & 5,912 annotated frames & 1,000 records \\
DVS-OUTLAB \cite{DVSOUTLAB2021} & Outdoor monitoring / activity analysis & object classes + environmental interferences & N/A & 47,878 labeled ROIs & several hours of event data; 3 sensors \\
DHP19 \cite{DHP192019} & 3D human pose estimation & 13 body joints & N/A & pose annotations & 33 movements $\times$ 10 repetitions $\times$ 17 subjects \\
NU-AIR \cite{NUAIRA2023} & Aerial traffic detection & pedestrian, vehicle & 2 & 93,204 boxes & 70.75 min \\
1Mpx \cite{1-M} & Automotive object detection & car, pedestrian, two-wheeler & 3 & $>$25 million boxes & $>$14 h driving data \\
MVSEC \cite{3dpp} & Stereo / flow / depth perception & N/A & N/A & geometric and pose annotations & multi-platform sequences \\
eTraM \cite{eTraM2024} & Traffic object detection & 8 traffic participant classes & 8 & $>$2 million boxes & 10 h \\
SEVD \cite{SEVD} & Synthetic multimodal traffic perception & car, truck, bus, bicycle, motorcycle, pedestrian & 6 & $\sim$9 million 2D/3D boxes & 58 h event data \\
TUMTrafEvent \cite{TUMTrafEvent} & Roadside traffic detection & 7 road-user classes & 7 & 50,496 2D boxes & 4,111 synchronized event-RGB frames \\
HARDVS2.0 \cite{HARDVS} & Human activity recognition & 300 daily activity classes & 300 & class labels & 107,646 paired sequences \\
DSEC \cite{DSEC} & Stereo depth / disparity / semantic perception & semantic classes & 11 / 19 & pixel-level geometric / semantic annotations & 53 sequences; $>$90 min \\
\hline
\end{tabular}
\end{adjustbox}
\end{sidewaystable}

\textbf{PEDRo}\footnote{https://github.com/SSIGPRO/PEDRo-Event-Based-Dataset} \cite{PEDRo2023} is a comprehensive dataset targeting human subjects in various environments and lighting conditions. The dataset contains manually annotated event data and grayscale imagery depicting varied actions by different individuals. A handheld camera was utilized during data collection, which, despite the integration of a buffering mechanism, introduced additional noise into the recordings. A distinctive characteristic of PEDRo is therefore its realistic viewpoint variation and motion-induced perturbation, making it more challenging than purely static-camera benchmarks. In addition, the dataset includes individuals aged between 20 and 70 and covers a broad spectrum of non-extreme weather conditions and diurnal variations, which makes it particularly valuable for studying person detection under diverse real-world sensing conditions.

\textbf{GEN1}\footnote{https://www.prophesee.ai/2020/01/24/prophesee-gen1-automotive-detection-dataset/} \cite{Prophesee2022}, the preeminent event-based dataset available for pedestrian detection in vehicular environments, employs GEN1 cameras mounted behind the windscreen of the car, alongside conventional grayscale cameras. Data collection spanned various French locales---from bustling urban centers and quiet towns to highways, rural stretches, and suburbs---across different diurnal and seasonal times and under varying meteorological conditions. More importantly, GEN1 has become one of the most influential automotive event-based benchmarks for object detection, since its acquisition setting closely matches realistic autonomous-driving perception.

\textbf{Henri's}\footnote{http://rpg.ifi.uzh.ch/e2vid} \cite{HSHDRV2022} dataset depicts a busy cityscape as seen from the front windscreen of a car in Zurich. The GEN3 camera and the rigid settings of a HUAWEI P20 smartphone were used to capture the footage, with both the event camera and frame camera recording at a resolution of 640$\times$480. While the data are not organized as a standard pedestrian-detection benchmark, the synchronized event and frame recordings document urban driving scenes under varying weather conditions and times of day, with a frequent focus on pedestrians and road users. Additionally, the dataset includes local weather data from Zurich, road condition information, and close-up lighting transitions, such as entering and exiting tunnels, thereby enriching the analysis of driving-related visual perception.

\textbf{PAFBenchmark}\footnote{https://github.com/CrystalMiaoshu/PAFBenchmark} \cite{Neuromo2019} covers three scenarios: pedestrian detection, motion detection, and fall detection, with the latter two documented within an open office setting. The pedestrian-detection component of the dataset encapsulates diverse scenarios including pedestrian overlap, occlusion, collision, and other common situations in traffic monitoring tasks. The action detection subset records 15 subjects performing 10 different actions, while the fall-detection subset similarly encompasses recordings of 15 subjects executing predefined motions such as falling, bending, tripping, and tying shoes. The dataset was captured using a fixed camera tripod and a laptop computer, and is therefore useful not only for pedestrian detection but also for broader human-centered event-based understanding.

\textbf{FJUPD}\footnote{https://github.com/fjcu-ee-islab/Spiking\_Converted\_YOLOv4} \cite{SpikeEvent2023} is an extension of the PAFBenchmark \cite{Neuromo2019} dataset for more detailed pedestrian-detection experiments, with a particular focus on outdoor scenes and low-light conditions. The data are divided into ``simple'' and ``complex'' categories based on luminosity and shadow dynamics, where complex backgrounds are characterized by pronounced light fluctuations and moving shadows. Compared with the original benchmark, its main value lies in providing more challenging lighting and background conditions, making it particularly useful for evaluating the robustness of event-based pedestrian detection under difficult illumination and shadow interference.

\textbf{DVS-OUTLAB}\footnote{http://dnt.kr.hsnr.de/DVS-OUTLAB/} \cite{DVSOUTLAB2021} uses a static acquisition method to capture activity within a fixed square area across a 2,800~m\textsuperscript{2} amusement park monitored by three fixed sensors. Positioned approximately 6 meters high with a 25-degree tilt toward the ground, the entire system operates on a self-sufficient solar energy storage system. Beyond long-duration outdoor event capture, the dataset also provides labeled regions of interest, including both object-related labels and environmental interferences such as rain and shadows. This makes DVS-OUTLAB particularly valuable for studying long-term outdoor event sensing under realistic environmental disturbances.

\textbf{DHP19}\footnote{https://github.com/uzh-rpg/event-based\_vision\_resources/} \cite{DHP192019} is the first dataset to use EVS for 3D human pose estimation, capturing the 3D positions of human joints through streams of events from multiple synchronized EVS cameras. This recording utilized four DAVIS cameras and the Vicon motion capture system, which consists of ten infrared cameras surrounding a motorized treadmill, thus facilitating varied movements by subjects. Although it is not a standard pedestrian-detection benchmark, DHP19 is highly relevant to EB-PD because it reveals how event streams encode fine-grained human motion structure, which is valuable for posture-sensitive pedestrian perception and behavior analysis.

\textbf{NU-AIR}\footnote{https://bit.ly/nuair-data} \cite{NUAIRA2023} consists of 70.75 minutes of event-camera footage captured by an EVS-equipped quadcopter, surveying diverse urban settings including crowds, various vehicles, and busy streetscapes. Annotations are provided at 30~Hz for pedestrians and vehicles, making it a valuable resource for aerial surveillance and event-based visual studies. Its main significance lies in extending EB-PD beyond ground-view driving scenarios to airborne perception settings.

\textbf{1Mpx}\footnote{https://www.prophesee.ai/category/dataset/} \cite{1-M} includes a diverse set of driving environments such as city streets, highways, and rural areas. A key feature of this dataset is the inclusion of very large-scale high-frequency annotations covering cars, pedestrians, and two-wheelers. The dataset was created using a novel automated labeling protocol that combines data from an event camera and a standard RGB camera. This extensive labeling and high resolution make it particularly valuable for developing and testing event-based object-detection systems under realistic driving conditions.

\textbf{MVSEC}\footnote{https://daniilidis-group.github.io/mvsec/} \cite{3dpp} is a large-scale multimodal dataset designed primarily for stereo depth estimation, visual odometry, and event-based 3D perception. It includes synchronized event streams, grayscale images, IMU measurements, LiDAR scans, and pose information collected from multiple platforms such as cars, motorcycles, handheld devices, and hexacopters. Rather than focusing on 2D pedestrian bounding boxes, MVSEC is especially valuable for geometry-aware event-based perception tasks. For EB-PD, its relevance lies in supporting pedestrian perception when integrated with depth, motion, and scene-structure understanding.

\textbf{eTraM}\footnote{https://eventbasedvision.github.io/eTraM} \cite{eTraM2024} offers fully event-based traffic footage recorded from a static overhead perspective using a Prophesee EVK4 HD event camera. Captured across diverse urban environments---including intersections, roadways, and local streets---the dataset features varying lighting and weather conditions and covers eight traffic participant classes such as pedestrians, vehicles, and micro-mobility. Its high resolution and overhead viewpoint make it a particularly valuable benchmark for event-based traffic monitoring and low-light object detection.

\textbf{SEVD}\footnote{https://eventbasedvision.github.io/SEVD/} \cite{SEVD} presents a large-scale fully synthetic multimodal traffic dataset including event streams, RGB, depth, optical flow, semantic masks, and instance segmentation. Captured in both ego-vehicle and static-camera configurations across diverse road types, SEVD supports multiple traffic-perception tasks and provides large-scale 2D/3D annotations across major traffic-participant categories. Its key value lies in its synthetic yet richly annotated multimodal nature, which makes it particularly useful for benchmarking multimodal event-based perception, large-scale supervised learning, and synthetic-to-real transfer.

\textbf{TUMTrafEvent}\footnote{https://innovation-mobility.com/en/project-providentia/a9-dataset/} \cite{TUMTrafEvent} provides synchronized event-based and RGB frames captured from a fixed roadside gantry overlooking a busy urban intersection. The dataset combines high-resolution RGB imagery with dense asynchronous event streams and manually verified 2D bounding boxes across multiple road-user categories. With its precise alignment and realistic roadside viewpoint, it serves as an important benchmark for event-RGB fusion in intelligent transportation systems and offers a useful complement to ego-vehicle datasets.

\textbf{HARDVS 2.0}\footnote{https://github.com/Event-AHU/HARDVS/tree/HARDVSv2} \cite{HARDVS} consists of paired RGB and event sequences recorded using a DAVIS346 sensor across indoor and outdoor scenes with diverse lighting and motion conditions. Captured at 30~FPS for RGB and microsecond resolution for events, the dataset includes 300 daily human activity categories under both static and dynamic camera setups. Although it is not a conventional pedestrian-detection dataset, HARDVS 2.0 is important because it expands event-based perception from object localization toward higher-level human-centered understanding, making it a comprehensive benchmark for multimodal human activity recognition in event-based vision.

\textbf{DSEC}\footnote{http://rpg.ifi.uzh.ch/dsec.html} \cite{DSEC} offers stereo event-based and frame-based driving data recorded in urban, suburban, and rural areas of Switzerland. It features two synchronized Prophesee Gen3.1 event cameras, stereo RGB cameras, LiDAR point clouds, and high-precision GPS. Unlike standard 2D detection datasets, DSEC primarily supports stereo depth estimation, optical flow, disparity, and related geometry-aware perception tasks, with dense pixel-level annotations and multiple driving sequences. Its relevance to EB-PD lies in the fact that pedestrian detection in autonomous driving is often only one component of a broader perception stack, and DSEC provides an important benchmark for understanding how event cameras contribute to geometry, motion, and scene-structure estimation.

\subsection{Evaluation Metrics}

The performance of EB-PD algorithms is commonly evaluated using three fundamental metrics: the log-average miss rate (MR), average precision (AP), and Intersection over Union (IoU) \cite{iou}. The MR quantifies detection failures by reflecting the proportion of ground-truth objects that are missed, whereas AP summarizes detection precision across different recall levels. IoU, in turn, measures the spatial overlap between predicted and ground-truth bounding boxes and serves as the basis for classifying predictions as true positives or false positives.

To determine true positives (TPs), false positives (FPs), and false negatives (FNs), a greedy matching strategy is typically adopted, as shown in Algorithm~\ref{algorithmic2}. Detections are first sorted according to confidence score, and each detection is then matched to the ground-truth instance with the highest IoU, provided that the overlap exceeds a predefined threshold. This matching procedure is central to the computation of both MR and AP.

Collectively, these metrics characterize different aspects of EB-PD performance. MR reflects detection sensitivity, AP captures the balance between precision and recall, and IoU measures spatial localization quality. In crowded scenes, where pedestrian overlap is frequent, IoU becomes particularly important because small localization errors can strongly affect whether a detection is counted as correct.

\begin{algorithm}
\caption{Greedy Matching for Detection Results}
\begin{algorithmic}[2]
\Require{$D$: List of detection bounding boxes with scores, $G$: List of ground-truth bounding boxes, $\alpha$: IoU threshold for matching detections to ground truths}
\Ensure{TP: List of True Positives, FP: List of False Positives, FN: List of False Negatives}
\Procedure{GreedyMatching}{$D$, $G$, $\alpha$}
    \State Sort $D$ by detection scores in descending order
    \State Initialize TP, FP, and FN as empty lists
    \For{each detection $d$ in $D$}
        \State Compute the IoU of $d$ with each ground-truth instance in $G$
        \If{the highest IoU is $\geq \alpha$}
            \State Mark $d$ as TP and remove the matched ground-truth from $G$
        \Else
            \State Mark $d$ as FP
        \EndIf
    \EndFor
    \State Mark all unmatched ground truths in $G$ as FN
    \State \Return TP, FP, FN
\EndProcedure
\end{algorithmic}
\label{algorithmic2}
\end{algorithm}

\textbf{Intersection over Union (IoU)} is analogous to the Jaccard index and measures the accuracy of spatial overlap between the predicted bounding box $B_d$ and the ground-truth bounding box $B_g$. It is defined as
\begin{equation}
    \mathrm{IoU}(B_d, B_g) = \frac{\mathrm{Area}(B_d \cap B_g)}{\mathrm{Area}(B_d \cup B_g)},
\end{equation}
where a higher IoU indicates more accurate localization. In most detection settings, a prediction is counted as a true positive only when the IoU exceeds a predefined threshold, commonly 0.5. The intersection area $|B_d \cap B_g|$ can be written as
\begin{align}
    & \max \bigl(0, \min(x_d^{\max}, x_g^{\max}) - \max(x_d^{\min}, x_g^{\min}) \bigr) \notag \\
    & \times \max \bigl(0, \min(y_d^{\max}, y_g^{\max}) - \max(y_d^{\min}, y_g^{\min}) \bigr),
\end{align}
and the union area $|B_d \cup B_g|$ is given by
\begin{align}
    A_{B_d} + A_{B_g} - |B_d \cap B_g|,
\end{align}
where $A_{B_d}$ and $A_{B_g}$ denote the areas of $B_d$ and $B_g$, respectively.

\textbf{Average Precision (AP)} is a standard metric derived from the precision--recall curve. It reflects detection quality across different recall levels and is commonly computed as
\begin{equation}
    AP = \sum_{k=1}^{n} \bigl(R(k) - R(k-1)\bigr) P(k),
\end{equation}
where $P(k)$ and $R(k)$ denote the precision and recall at the $k$-th threshold, respectively. Precision is the ratio of true positive detections to all detections, whereas recall is the ratio of true positive detections to all actual positives.

\textbf{Miss Rate (MR)} is particularly useful for imbalanced detection tasks, where background samples substantially outnumber object instances. It is defined as
\begin{equation}
    MR = 1 - TPR = 1 - \frac{TP}{TP + FN},
\end{equation}
where $TPR$ is the true positive rate, $TP$ denotes the number of true positives, and $FN$ denotes the number of false negatives. In pedestrian detection, the log-average miss rate is often computed over a predefined range of false positives per image on a logarithmic scale, typically from $10^{-2}$ to $10^{0}$.
\section{OPEN CHALLENGES AND FUTURE PROSPECTS}

\subsection{Real-World Challenges and Future Perspectives}

Although EB-PD has advanced substantially in recent years, its large-scale real-world deployment remains constrained by several intertwined challenges. These challenges are not limited to algorithmic performance, but also involve hardware cost, dataset standardization, evaluation methodology, and the lack of sufficiently mature event-native perception models. Consequently, future progress in EB-PD depends not only on improving detection accuracy, but also on strengthening the practical foundations required for robust and scalable deployment.

\subsubsection{Hardware cost, industrialization, and deployment readiness}

A primary obstacle to the widespread adoption of event-based cameras is the current difficulty of industrial-scale deployment. Many event-based systems still rely heavily on Field-Programmable Gate Arrays (FPGAs) as their main control units \cite{DAVIS1, DAVIS2}. While such hardware offers flexibility for prototyping and sensor interfacing, it also significantly increases system cost, complexity, and integration difficulty. These factors hinder large-scale commercialization, especially in cost-sensitive sectors such as mainstream automotive systems, intelligent roadside infrastructure, and consumer-grade edge devices.

This hardware issue is closely related to the broader problem of deployment readiness. Despite the attractive theoretical advantages of EVS, including low latency, high dynamic range, and sparse data output, large-scale real-world studies remain limited. Current evidence remains insufficient to determine when and under what conditions event-based pedestrian detection can consistently outperform or reliably complement mature frame-based systems across diverse autonomous-driving environments. Therefore, broader real-world validation, long-term field trials, and hardware-software co-design remain essential before EB-PD can become a dependable industrial solution.

\subsubsection{Dataset standardization and evaluation limitations}

Another major challenge lies in the lack of standardized large-scale datasets and unified evaluation protocols specifically tailored to event-based pedestrian detection. Existing datasets vary substantially in sensor type, annotation quality, environmental diversity, class definitions, and data format, which complicates fair comparison across methods. In many cases, models are still validated on self-collected datasets or on event datasets originally developed for tasks beyond pedestrian detection. Although such datasets are valuable, they often limit reproducibility and weaken the comparability of published results.

In parallel, the evaluation of EB-PD still depends largely on metrics inherited from frame-based object detection. While standard metrics such as AP, MR, and IoU remain useful, they do not fully characterize key properties of event-based sensing, such as temporal fidelity, asynchronous responsiveness, and robustness under extreme dynamic illumination. As a result, there is still no complete evaluation framework specifically designed for EB-PD. Future work should therefore focus on both large-scale standardized benchmarks and more task-relevant evaluation criteria that better reflect the unique sensing characteristics of event cameras.

\subsubsection{Model specialization and future research directions}

A further limitation of current EB-PD research is the insufficient development of models specifically designed for event data. Many existing approaches still process event streams through frame-like conversion or rely on conventional backbones originally developed for RGB imagery. While these strategies are practical and often effective, they do not fully exploit the fine-grained temporal structure and sparse asynchronous nature of event streams. This is particularly limiting from a three-dimensional and motion-aware perspective, where event data potentially contain richer cues than those captured by frame-based representations.

To address this gap, future research should move beyond adapting existing vision models and instead develop architectures that are intrinsically compatible with event-based sensing. This includes models that can more effectively mine spatio-temporal structure, exploit sparse asynchronous patterns, and jointly optimize sensing, preprocessing, and inference. In addition, progress will likely require stronger collaboration between academia and industry, not only for model design, but also for sensor engineering, standardization, and large-scale validation. In this sense, the future of EB-PD depends on a transition from proof-of-concept systems to integrated, benchmarked, and deployment-oriented event-native perception pipelines.

\subsection{Research Trends and Future}

Looking forward, the future of pedestrian detection with event-based sensors is likely to be shaped by a convergence of advances in hardware, adaptive perception, multimodal learning, and responsible deployment. These directions are closely connected: improvements in sensing hardware affect algorithm design, while progress in representation learning and system integration will determine whether event cameras can transition from niche experimental devices to widely usable perception components.

\subsubsection{Edge intelligence and hardware-software co-design}

One of the most promising research trends is the integration of event cameras with edge-intelligent and sense-compute convergence technologies. Because event-based cameras naturally produce sparse outputs and operate with low latency, they are well suited to scenarios requiring efficient on-device perception. In principle, tighter integration between sensing and computation could enable real-time processing with lower energy consumption and reduced communication overhead, which is particularly attractive for autonomous vehicles, wearable devices, drones, and IoT systems.

At the same time, this trend is strongly related to the need for cost reduction and industrial scalability. Future progress may depend partly on reducing reliance on expensive and relatively inflexible FPGA-dominated solutions and developing more integrated, application-specific hardware. Such developments could make event-based sensing more accessible and commercially viable, while also encouraging the design of algorithms that are more tightly matched to deployment constraints.

\subsubsection{Adaptive perception and multimodal event intelligence}

Another important direction is the development of event-camera systems with stronger environmental adaptability. In practical applications, the usefulness of event sensors depends not only on their temporal precision, but also on their ability to remain stable across changing illumination, cluttered backgrounds, and complex motion patterns. Continued progress in denoising, adaptive thresholding, bias tuning, sensor calibration, and self-adaptive representation learning may substantially improve the reliability of event-based perception under real-world variability.

This direction also naturally connects with multimodal learning. Event cameras alone provide motion-sensitive and temporally precise information, but often lack the dense appearance cues available in frame-based sensing. As a result, future systems are likely to rely increasingly on multimodal fusion, combining events with RGB images, depth, LiDAR, or inertial signals. Such integration may enable richer perception pipelines in which event streams contribute temporal sensitivity while complementary modalities provide spatial and semantic completeness.

\subsubsection{Broader applications and responsible deployment}

Although autonomous driving remains one of the most important application domains for EB-PD, the scope of future deployment is much broader. Event-based pedestrian perception may also play an important role in smart-city infrastructure, roadside traffic monitoring, public safety, human behavior analysis, wearable assistance, and robotic perception in visually challenging environments. As event-based technology matures, its applications are likely to expand from specialized experimental settings toward more diverse human-centered perception tasks.

At the same time, broader deployment will inevitably raise regulatory and ethical questions. Since many applications involve surveillance, monitoring, or large-scale urban sensing, future research must address privacy, responsible data governance, and compliance with relevant legal frameworks. Therefore, the long-term development of EB-PD should not be viewed solely as a technical trajectory. Its success will also depend on whether advances in sensing and algorithms can be aligned with social acceptance, regulatory requirements, and practical deployment needs.

In summary, the future of EB-PD will be shaped not only by improvements in detection algorithms, but also by progress in sensor design, system integration, benchmark standardization, adaptive perception, and ethical deployment. Continued innovation across these dimensions will determine whether event-based pedestrian detection can evolve from a promising research topic into a robust and widely adopted real-world technology.

\section{Conclusion}
\label{sec:conclusion}

This review has systematically examined the development of event-based vision sensing for pedestrian detection, with particular emphasis on its relevance to autonomous driving, intelligent transportation, and intelligent surveillance. By reviewing the literature from sensing principles to downstream detection pipelines, we have outlined the technological evolution of EVS and highlighted its key advantages over conventional frame-based imaging, including low latency, high temporal resolution, sparse data output, and strong robustness under challenging illumination.

We further reviewed the major methodological components of EB-PD, including sensing and processing paradigms, preprocessing strategies, feature representations, detection models, datasets, and evaluation metrics. In particular, we emphasized that existing EB-PD methods should not be understood merely as separate implementations, but rather as different design choices along shared trade-offs involving temporal fidelity, detection accuracy, computational efficiency, and deployment complexity. From this perspective, direct event-stream methods, event-to-frame methods, and event-frame fusion approaches each occupy different but complementary positions within the current research landscape.

Despite recent progress, several important challenges remain. These include the scarcity of standardized large-scale benchmarks tailored to EB-PD, the continued reliance on frame-based evaluation conventions, the limited maturity of event-native detection models, and the difficulty of translating promising laboratory results into robust real-world systems. Future progress will likely depend on advances in event-specific representation learning, hardware-software co-design, multimodal fusion, adaptive perception, and deployment-oriented evaluation.

Overall, event-based vision offers a compelling direction for next-generation pedestrian perception. Although the field is still evolving, continued progress in sensing technology, model design, and benchmark development is likely to make EB-PD an increasingly important component of future intelligent perception systems.

\section*{Acknowledgment}
This work was supported by the National Key R\&D Program of China (Grant No. 2023YFB4503000), the Hefei Key Technology R\&D ``List and Command'' Project (Grant No. 2023SGJ035), the HFIPS Director's Fund (Grant No. YZJJKX202401), and the Hefei Municipal Natural Science Foundation (Grant No. HZR2414).

\bibliographystyle{elsarticle-num} 
\bibliography{reference}

\end{document}